\documentclass[journal]{IEEEtran}

\usepackage[utf8]{inputenc}
\usepackage[T1]{fontenc} 
\usepackage{cite}
\usepackage{amsmath,amsfonts,amssymb,amsthm} 
\usepackage{bm}
\usepackage{mathrsfs}
\usepackage{latexsym}

\usepackage{graphicx}
\usepackage[caption=false,font=normalsize,labelfont=sf,textfont=sf]{subfig} 
\usepackage{stfloats}
\usepackage{float}
\usepackage{epstopdf} 

\usepackage{array}
\usepackage{booktabs}
\usepackage{multirow}
\usepackage{makecell}
\usepackage{tabularx}
\usepackage{longtable}
\usepackage{threeparttable}
\usepackage{algorithm}
\usepackage{algorithmic}

\usepackage{url}
\usepackage{verbatim}
\usepackage{indentfirst}
\usepackage{ragged2e}
\usepackage[acronym]{glossaries}
\usepackage{rotating}
\usepackage{balance}
\usepackage[hyphenbreaks]{breakurl}
\usepackage[cmyk]{xcolor} 
\usepackage{hyperref}
\usepackage{threeparttable}
\usepackage{balance}
\newtheorem{property}{Property}

\theoremstyle{plain}

\theoremstyle{definition}

\theoremstyle{remark}

\begin{document}
	
	\title{WLC: Weber-Inspired Local Contrast Metric for Low-Altitude Image Fusion}

	\author{Cong Wang,~\IEEEmembership{Member,~IEEE}, Yufeng Xie, and Xuelong Li,~\IEEEmembership{Fellow,~IEEE}
		\thanks{This work was supported in part by the National Natural Science Foundation of China under Grant 62476219, in part by the National Key R\&D Program of Shaanxi under Grant 2024CY2-GJHX-54, in part by the Young Talent Fund of Association for Science and Technology in Shaanxi, China under Grant 20230140, and in part by the Fundamental Funds for the Central Universities.}
		\thanks{C. Wang is with the School of Mathematics and Statistics, Northwestern Polytechnical University, Xi'an, China (e-mail: congwang0705@nwpu.edu.cn).}
		\thanks{Y. Xie is with the School of Mathematics and Statistics, Northwestern Polytechnical University, Xi'an, China (e-main: yufengxie@mail.nwpu.edu.cn).}
		\thanks{X. Li is with the Institute of Artificial Intelligence (TeleAI) of China Telecom, China (e-mail: xuelong\_li@ieee.org).}
		}

	
	\maketitle
	
	\begin{abstract}
		Infrared and visible image fusion is a pivotal technology in low-altitude Unmanned Aerial Vehicle (UAV) reconnaissance missions, enabling robust target detection and tracking by integrating thermal saliency with environmental textures. 
		However, the advancement of fusion algorithms is hindered by a critical evaluation bottleneck. 
		In this paper, we identify a systematic failure in traditional no-reference metrics (specifically Statistics-based and Gradient-based metrics) within complex low-light environments, termed as ``Noise Trap''. 
		It is mathematically proven that these metrics are positively correlated with high-frequency sensor noise, paradoxically assigning higher scores to degraded images and misguiding algorithm optimization. 
		To resolve this dilemma, this paper proposes the Weber-inspired Local Contrast (WLC) metric. 
		Grounded in the psychophysical principle of Weber's Law, WLC shifts the evaluation paradigm from global statistical distribution to local semantic contrast. By leveraging infrared priors, it effectively decouples target saliency from global background noise. 
		Extensive experiments on the DroneVehicle dataset demonstrate that WLC exhibits high ``Semantic Discriminability'' in distinguishing thermal targets from background clutter. Furthermore, it achieves remarkable computational efficiency, thus, establishing itself as a reliable and real-time standard for intelligent UAV systems.
	\end{abstract}
	
	\begin{IEEEkeywords}
		Image Fusion, Evaluation Metric, Low-Altitude Reconnaissance, Weber's Law.
	\end{IEEEkeywords}

	\section{Introduction}
	
	\IEEEPARstart{M}{ulti-modal} perception constitutes the bedrock of situational awareness in modern intelligent surveillance systems. In the specific context of low-altitude Unmanned Aerial Vehicle reconnaissance, the visual environment is fraught with challenges, ranging from extreme illumination changes to electronic noise interference. Single-sensor systems often fail to provide comprehensive scene descriptions under such conditions. Infrared sensors, while robust in capturing the thermal radiation of salient targets regardless of lighting \cite{4ir}, suffer from low spatial resolution and a lack of textural definition. Conversely, visible sensors excel at preserving environmental context and detailed textures \cite{5vis} but are severely compromised in low-light scenarios. To transcend these physical limitations, Infrared and Visible Image Fusion (IVIF) has emerged as a pivotal technology \cite{r1,r2}. By synergizing thermal saliency with background texture into a composite representation, IVIF provides high-quality data support that is critical for downstream decision-making tasks, including target detection \cite{jc1}, robust tracking \cite{zz1}, image segmentation \cite{ref20}, and spectral reconstruction \cite{ref26}.
	
	The evolution of IVIF algorithms over the past decade reflects a paradigm shift from manual signal processing to data-driven representation learning. Early methodologies were dominated by Multi-Scale Transforms (MST) and sparse representation, which decomposed images into base and detail layers for rule-based fusion. With the advent of deep learning, the field has bifurcated into several dominant streams: Auto-Encoder (AE) frameworks for feature extraction \cite{joint6}, Convolutional Neural Networks (CNNs) for end-to-end mapping \cite{CNN7}, and Generative Adversarial Networks (GANs) for distribution alignment \cite{fusiongan}. More recently, Transformer-based architectures have been introduced to capture long-range dependencies, further pushing the boundaries of visual generation quality \cite{crossfuse,swinfusion}.
	Despite these algorithmic strides, a critical asymmetry persists in the field: \textit{While generation models have become increasingly sophisticated, the evaluation mechanisms used to benchmark them remain archaic.} The community continues to rely heavily on general-purpose No-Reference metrics, such as Entropy (EN) and Average Gradient (AG), which were originally designed for undistorted natural images rather than fused imagery in degraded sensing environments \cite{MemoryFusion,T2EA,FreqGAN}.
	
	This study is motivated by a fundamental misalignment between traditional evaluation protocols and the specific requirements of low-altitude reconnaissance. Traditional metrics operate on a statistical hypothesis: \textit{``higher pixel complexity or sharper gradients indicate better image quality''.} However, in low-altitude and remote sensing scenarios, the visual environment poses significant challenges due to occlusions and scale variations \cite{ref34}. In such scenarios, visible images are plagued by severe high-frequency sensor noise. Paradoxically, existing metrics misinterpret this noise as ``rich textural information''. This phenomenon is termed as ``Noise Trap''.
	
	\begin{figure*}[htbp]
		\centering
		\includegraphics[width=0.9\linewidth]{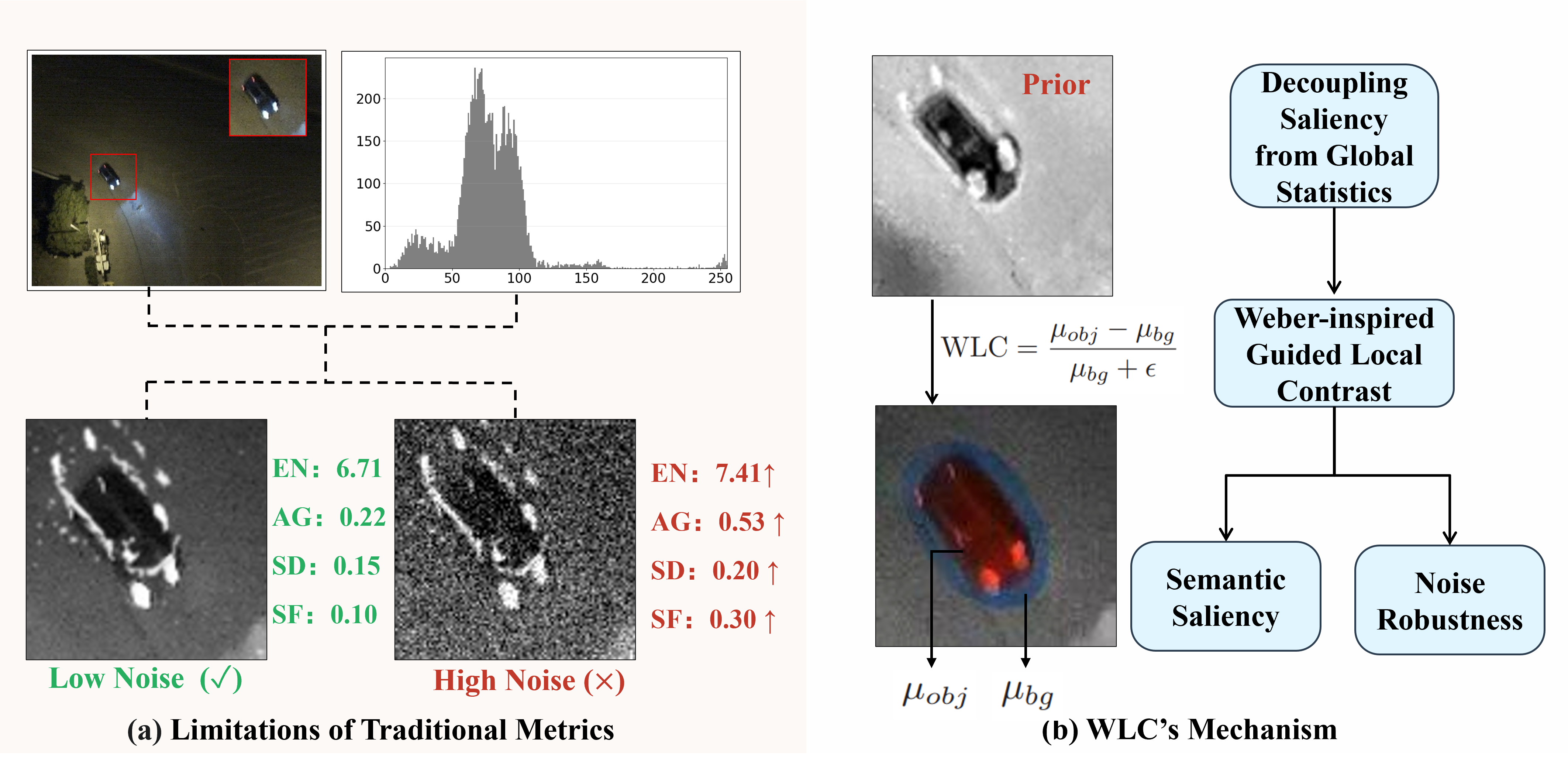}
		\caption{The paradigm shift in fusion evaluation: From Global Statistics to Local Semantics. 
		(a) Limitations of Traditional Metrics: Traditional metrics (e.g., AG, EN, SD, and SF) rely on global statistical distributions (represented by the histogram). Consequently, they misinterpret high-frequency sensor noise as valid information, paradoxically assigning higher scores to the ``High Noise'' image (e.g., AG surges from 0.22 to 0.53). 
		(b) WLC's Mechanism: Grounded in Weber's Law, the metric decouples saliency from global noise. By utilizing the infrared prior to define the semantic target $\mu_{obj}$ and its local background $\mu_{bg}$, WLC measures the relative contrast, ensuring robust evaluation aligned with human perception.}
		\label{fig:1}
	\end{figure*}
	
	To visually illustrate this conflict, Fig. \ref{fig:1} contrasts the failure mechanism of traditional metrics with the proposed physical paradigm. As shown in Fig. \ref{fig:1}(a), traditional metrics operate under a global statistical paradigm. They aggregate pixel-level information (e.g., gradients or histogram entropy) across the entire image. When the scene is corrupted by severe sensor noise, the global statistical complexity increases. Consequently, metrics like AG and EN fall into the aforementioned ``Noise Trap'', erroneously identifying the noisy, low-quality image as superior. This contradicts the visual reality where the target is barely distinguishable. Crucially, current investigation reveals that this failure is not isolated to specific metrics but is a systematic defect across metric families: Gradient-based metrics mathematically equate high-frequency noise with texture gradients, while Statistics-based metrics mistake noise histogram spreading for information gain.
	
	This evaluation failure misguides the optimization of modern fusion networks. To mitigate this, the research community has witnessed a shift towards task-driven fusion. Recent prominent works, such as SeAFusion \cite{SeAfuse} and TarDAL \cite{TarDAL}, integrate high-level semantic constraints into the fusion framework. Consequently, the performance of downstream object detection, quantified by mAP\cite{Lin2014COCO}, has become a \textit{de facto} standard for assessing the practical utility of fused images. Conceptually, mAP computes the area under the Precision-Recall curve across various Intersection over Union (IoU) thresholds, providing a rigorous assessment of both localization and classification capabilities of a detector. However, relying solely on mAP for evaluation presents significant dilemmas: it is computationally prohibitive, requires large-scale annotated datasets, and involves heavy inference models. This makes mAP impractical for real-time, unsupervised quality assessment on resource-constrained edge devices.
	
	To resolve this dilemma, this paper aims to establish a lightweight, and unsupervised evaluation standard that aligns with machine perception while remaining immune to noise interference. The technical approach departs from pure statistical measurement, turning instead to psychophysical principles. As a result, Weber-Inspired Local Contrast (WLC) is proposed. As visually conceptualized in Fig. \ref{fig:1}(b), unlike traditional metrics that measure global pixel distributions, it embeds the ``thermal saliency prior'' to focus exclusively on local semantic contrast. Specifically, recognizing that small target detection requires local saliency rather than just global contrast \cite{target}, morphological operations are leveraged to construct a topological model consisting of the target's thermal core and its immediate local neighborhood. By calculating the relative luminance gain between these two regions, WLC effectively decouples target saliency from global background noise. This design ensures that the metric selectively rewards the visibility of small targets while imposing a heavy penalty on clutter. Furthermore, thanks to its linear computational complexity, WLC supports real-time evaluation.
	
	The originality of this work lies in the systematic adaptation of Weber's Law\cite{weibo-Psychophysics} to the domain of low-altitude image fusion evaluation. While Weber's Law is a foundational principle in psychophysics describing human contrast sensitivity, its application in benchmarking IVIF algorithms has remained unexplored. This gap is bridged by mathematically formulating the physical law into a fusion metric. It is demonstrated that the ``relative contrast'' defined by Weber's Law is a far more reliable indicator of fusion quality than the ``absolute intensity'' measured by traditional metrics, particularly in scenarios where the background is dominated by noise. This represents a methodological shift from ``signal-level statistics'' to ``semantic-level perception''.
	
	The main contributions of this paper are summarized as follows:
	\begin{itemize}
		\item It provides a systematic analysis of the failure mechanisms of Gradient-based and Statistics-based metrics in low-light scenarios. Through mathematical derivation (e.g., Rayleigh distribution of noise gradients) and empirical evidence, it demonstrates that these metrics exhibit a positive correlation with noise intensity, leading to ``Noise Trap'' and erroneous quality assessments.
		
		\item It proposes WLC metric grounded in Weber's Law. By utilizing infrared priors to locate salient targets and employing morphological operations to model the topology of the immediate local background, it effectively decouples target saliency from global background statistics. Large-scale statistical analysis reveals that WLC possesses high ``Semantic Discriminability'', whereas traditional metrics exhibit ``Semantic Numbness''.
		
		\item It establishes a comprehensive validation framework based on synthetic degradation and real-world testing. Extensive experiments on the DroneVehicle \cite{sun2020drone} dataset verify that WLC achieves superior robustness compared to traditional metrics and exhibits a positive correlation with downstream detection performance. Furthermore, it demonstrates remarkable efficiency (approximately 3 times faster than AG), making it ideal for real-time Unmanned Aerial Vehicle applications.
	\end{itemize}
	
	The remainder of this paper is organized as follows. Section \ref{sec:related} reviews related work on fusion algorithms and evaluation methodologies. Section \ref{sec:method} details the mathematical formulation of WLC and its theoretical basis. Section \ref{sec:validation} presents the experimental validation, including synthetic stress tests, real-world statistical analysis, and computational complexity benchmarking. Finally, Section \ref{sec:conclusion} concludes the paper and discusses future directions.
	
	\section{Related Work}
	\label{sec:related}
	
	This section scrutinizes the trajectory of infrared and visible image fusion algorithms and the parallel evolution of objective evaluation metrics. Specifically, it dissects the limitations of existing metrics within low-altitude reconnaissance scenarios and elucidates the theoretical motivation underpinning the proposed WLC metric.
	
	\subsection{Infrared and Visible Image Fusion Algorithms}
	The landscape of image fusion methodologies has undergone a paradigm shift from manual signal processing to data-driven deep learning over the past decades. Early approaches were predominantly anchored in Multi-Scale Transforms, utilizing techniques such as Wavelet Transform \cite{Wave} and Laplacian Pyramid \cite{Pyramid} to decompose images into base and detail layers for rule-based fusion. While computationally efficient, these methods often grapple with inadequate feature representation in complex, non-linear scenarios.
	
	The advent of deep learning has revolutionized this domain. Auto-Encoder frameworks, exemplified by DenseFuse \cite{densefuse}, introduced learnable feature extraction and reconstruction, offering a significant leap in texture preservation. Subsequently, Generative Adversarial Networks, such as FusionGAN \cite{fusiongan}, modeled the fusion process as a minimax game, compelling the fused image to retain the thermal intensity of infrared targets. More recently, the focus has pivoted towards capturing long-range dependencies, with Transformer-based architectures like SwinFusion \cite{swinfusion} setting new benchmarks for visual quality. 
	Despite these algorithmic strides, a critical asymmetry persists: \textit{while generation models have become increasingly sophisticated, the quantitative metrics used to benchmark them have remained largely stagnant.} This lag creates a misalignment between generation quality and evaluation standards, potentially hindering further progress.
	
	\subsection{Objective Evaluation Metrics}
	Given the unsupervised nature of image fusion, which typically lacks Ground Truth, objective evaluation relies heavily on No-Reference metrics. Recently, structural-equation-modeling-based approaches have been introduced to construct comprehensive indicator systems for image quality assessment \cite{ref28}, which underscores that evaluating image fusion requires a multi-dimensional perspective rather than reliance on single statistical indicators. While existing metrics can be broadly categorized into five types \cite{deep}, the discussion herein focuses on the three primary categories most relevant to reconnaissance tasks.
	
	\subsubsection{Information Theory-based Metrics}
	Metrics such as Entropy (EN) \cite{EN} and Mutual Information (MI) \cite{MI} quantify the information content based on statistical distributions. These metrics predicate on the assumption that ``higher complexity equals more information.'' However, in low-altitude night scenarios, high-frequency sensor noise significantly flattens the pixel histogram, increasing statistical randomness. Consequently, EN tends to interpret this noise-induced uncertainty as valid information, assigning paradoxically high scores to low-quality, noisy images.
	
	\subsubsection{Image Feature-based Metrics}
	Spatial Frequency (SF) \cite{SF} and Average Gradient (AG) \cite{AG} assess texture richness and image clarity by calculating gradient magnitudes. Although effective for noise-free images, these metrics exhibit pathological sensitivity to high-frequency components. In low-light environments characterized by low Signal-to-Noise Ratio, background noise often manifests as strong gradients. Therefore, AG and SF inherently fall into the ``Noise Trap,'' failing to distinguish between valid structural edges and random sensor noise.
	
	\subsubsection{Perception-based Metrics}
	Metrics like Visual Information Fidelity (VIF) \cite{VIF} attempt to model the Human Visual System to assess image naturalness and distortion. While more robust than simple statistical metrics, they typically treat all pixels with equal importance or focus on global structural consistency. They fail to account for the specific semantic requirement of reconnaissance tasks—the relative saliency of small thermal targets against a dark, cluttered background.
	
	\subsection{Task-Driven Evaluation and the Metric Gap}
	Recognizing the semantic deficit of general-purpose metrics, recent research has shifted towards task-driven evaluation. State-of-the-art methods, such as SeAFusion \cite{SeAfuse} and TarDAL \cite{TarDAL}, explicitly incorporate high-level vision tasks (e.g., object detection) into the fusion framework. With the rise of these task-oriented methods, detection accuracy (mAP) has emerged as a rigorous, albeit expensive, evaluation standard. However, while mAP accurately reflects machine perception utility, it presents significant practical challenges: it is computationally prohibitive, requires large-scale annotated datasets, and involves heavy inference models, rendering it unsuitable for unsupervised, real-time quality assessment.
	
	The proposed WLC metric reconciles this dichotomy. Unlike traditional metrics that are blind to semantics, WLC integrates the ``thermal saliency prior'' inspired by task-driven methodologies. Simultaneously, unlike mAP, WLC maintains the unsupervised nature and computational efficiency required for edge deployment (as discussed in Section \ref{sec:validation}). By grounding the evaluation in Weber's Law, WLC offers a mathematically sound and physically interpretable standard specifically tailored for low-altitude small target detection.
	
	\section{Proposed Methodology}
	\label{sec:method}
	
	This section details the theoretical framework of the proposed evaluation metric. First, this section provides a mathematical derivation explaining the failure mechanism of traditional metrics in noisy environments, categorizing them into statistics-based and gradient-based families. Subsequently, it introduces WLC, elaborating on its psychophysical foundation, algorithmic formulation, and theoretical properties.
	
	\subsection{Theoretical Analysis of ``Noise Trap''}
	Existing no-reference evaluation metrics typically rely on information theory. They assume that high-frequency changes represent valid information. However, this assumption is flawed in low-altitude night scenarios. 
	
	To rigorously explain the failure of these metrics, we analyze the statistical behavior of traditional metrics under Gaussian noise. 
	This paper assumes that each pixel is independently and identically distributed. Then, the observed visible image $\mathbf{V}$ is modeled as a superposition of the ideal scene signal $\mathbf{I}$ and additive sensor noise $\mathbf{N}$\cite{congwang}:
	\begin{equation}
		\label{eq: model}
		\mathbf{V}(x,y) = \mathbf{I}(x,y) + \mathbf{N}(x,y)
	\end{equation}
	where $(x,y)$ is the pixel coordinate, $\mathbf{N}$ typically follows a Normal distribution $\mathcal{N}(0, \sigma^2)$ caused by ISO sensitivity \cite{Digital}.
	
	\subsubsection{Statistics-based Family (Entropy and Standard Deviation)}
	This category of metrics measures the global dispersion or uncertainty of pixel intensities.
	
	\paragraph{Standard Deviation (SD)}
	It reflects the dispersion of pixel values in a fused image F $\subset \mathbb{R}^{M \times N}$ relative to the mean intensity. It is mathematically defined as:
	\begin{equation}
		\mathrm{SD} = \sqrt{\frac{1}{MN} \sum_{x=1}^{M} \sum_{y=1}^{N} (\mathbf{F}(x,y) - \mu)^2}
	\end{equation}
	where $\mu$ is the mean value of the image. $M$ and $N$ are the size of the image, representing the length and width.  
	Under this additive model (\ref{eq: model}), the variance of the observed image follows the additivity property: 
	\begin{equation}
		Var(\mathbf{V}) = Var(\mathbf{I}) + Var(\mathbf{N}) = Var(\mathbf{I}) + \sigma^2
	\end{equation}
	Consequently, the SD score scales with the noise level: 
	\begin{equation}
		\mathrm{SD}(\mathbf{V}) = \sqrt{Var(\mathbf{I}) + \sigma^2}
	\end{equation}
	As noise $\sigma$ increases, SD monotonically rises.
	
	\paragraph{Entropy (EN)}
	It quantifies the average information content. For an image $\mathbf{F}$, it is defined as
	\begin{equation}
		\mathrm{EN} = - \sum_{l=0}^{L-1} p_l \log_2 p_l
	\end{equation}
	where $L$ is the number of gray levels, and $p_l$ is the probability of the occurrence of gray level $l$.
	
	According to information theory, a completely random noise signal possesses maximum entropy. In night images, ISO noise flattens the histogram distribution. This artificially inflates the EN score. 
	
	When subjected to additive Gaussian noise $\mathbf{N} \sim \mathcal{N}(0, \sigma^2)$, the pixel distribution $P(x)$ spreads out, approximating a Gaussian distribution $\mathcal{N}(c, \sigma^2)$. 
	According to information theory\cite{Probability}, the differential entropy $H$ of a Gaussian variable is maximized for a given variance and is derived as
	\begin{equation}
		\label{eg:en}
		\mathrm{EN} = H = \frac{1}{2} \log_2 (2\pi e \sigma^2) = \log_2 \sigma + \frac{1}{2} \log_2 (2\pi e)
	\end{equation}
	where $e$ is the natural constant.
	
	In (\ref{eg:en}), it indicates a relationship, i.e., the EN score scales with $\log_2 \sigma$.
	As sensor noise $\sigma$ increases, the pixel histogram flattens, representing higher statistical ``uncertainty''. Traditional metrics misinterpret this noise-induced uncertainty as ``information richness'', causing EN scores to rise initially.
	
	\subsubsection{ Gradient-based Family (Average Gradient and Spatial Frequency)}
	This category quantifies texture clarity by calculating gradient magnitudes or frequency responses.
	
	\paragraph{Average Gradient (AG)}
	It is used to quantify the minute details and texture clarity of an image $\mathbf{F}$:
	\begin{equation}
		\label{eg:AG}
		\mathrm{AG} = \frac{1}{MN} \sum_{x=1}^{M} \sum_{y=1}^{N} \sqrt{\frac{\nabla_x \mathbf{F}(x,y)^2 + \nabla_y \mathbf{F}(x,y)^2}{2}}
	\end{equation}
	where $\nabla_x$ and $\nabla_y$ represent the differences in the horizontal and vertical directions, respectively.
	
	Let $\nabla$ denote the gradient operator. The gradient of the observed image is:
	\begin{equation}
		\nabla \mathbf{V} = \nabla \mathbf{I} + \nabla \mathbf{N}
	\end{equation}
	
	Since this paper investigates the relationship between AG and noise, $\nabla \mathbf{I}$ can be considered a constant. The observed gradient is thus dominated by the noise gradients. By (\ref{eg:AG}), specifically, $\nabla_x \mathbf{N}(x,y), \nabla_y \mathbf{N}(x,y) \sim \mathcal{N}(0, \sigma^2)$.
	
	According to the probability theory and (\ref{eg:AG}), the magnitude $\mathbf{R}$ of a vector composed of two zero-mean Gaussian components follows a Rayleigh Distribution\cite{Raliy}. 
	\begin{equation}
		\mathbf{R} \triangleq \sqrt{\nabla_x \mathbf{N}(x,y)^2 + \nabla_y \mathbf{N}(x,y)^2}
	\end{equation}
	
	The scale parameter of this Rayleigh distribution is the standard deviation of the gradients, i.e., $s = \sqrt{2}\sigma$.
	
	The expected value of the AG in noise-dominated regions is therefore derived from the expectation of the Rayleigh distribution:
	\begin{equation}
		\label{eq:EM}
		\mathbb{E}[\mathrm{AG}] \approx \mathbb{E}[\mathbb{R}] = s \sqrt{\frac{\pi}{2}} = (\sqrt{2}\sigma) \cdot \sqrt{\frac{\pi}{2}} = \sqrt{\pi} \sigma
	\end{equation}
	where $\mathbb{E}$ is the mathematical expectation.
	
	In (\ref{eq:EM}), it reveals a strict linear relationship, i.e., the AG score scales proportionally with $\sqrt{\pi} \sigma$.
	Therefore, as sensor noise $\sigma$ increases, AG monotonically increases, misleadingly rewarding noisy images with higher quality scores. 
	
	\paragraph{Spatial Frequency (SF)}
	It measures the overall activity level of an image based on row frequency (RF) and column frequency (CF).
	It is defined as:
	\begin{align}
		\mathrm{SF} &= \sqrt{\mathrm{RF}^2 + \mathrm{CF}^2} \\
		\text{with} \quad \mathrm{RF} &= \sqrt{\frac{1}{MN}\sum (\mathbf{F}(x,y)-\mathbf{F}(x,y-1))^2} \nonumber \\
		\phantom{\text{with}} \quad \mathrm{CF} &= \sqrt{\frac{1}{MN}\sum (\mathbf{F}(x,y)-\mathbf{F}(x-1,y))^2} \nonumber
	\end{align}
	
	Mathematically speaking, the terms inside the square are equivalent to AG. Consequently, SF shares an isomorphic statistical behavior with AG under Gaussian noise. As $\sigma$ increases, the squared differences increase, causing SF to scale linearly with the noise level. In summary, as sensor noise $\sigma$ increases, both the Statistics-based metrics and the Gradient-based metrics monotonically increase, misleadingly rewarding noisy images with higher quality scores. This phenomenon is termed as ``Noise Trap''.
	
	\subsubsection{Weber's Law}
	While a whisper is easily audible in a quiet room, a shout might go unheard in a noisy environment. This captures the essence of Weber's Law\cite{weibo-Psychophysics}. It hypothesizes that the ratio of the just-noticeable difference in a stimulus $\Delta I$ to the background level of the stimulus $I$ is a constant.
	\begin{equation}
		k = \frac{\Delta I}{I}
	\end{equation}
	where $k$ is known as the Weber fraction, implying that despite variations in $\Delta I$ and $I$, this ratio remains constant.
	
	It suggests that perception depends on relative change, not absolute intensity. Similar concepts are successfully applied in texture segmentation\cite{segment-weibo}, face recognition \cite{face-weibo}, brain images\cite{brain-weibo}, tactile feedback\cite{feedback-weibo}, echo processing\cite{echo-weibo}
	
	In the context of low-altitude reconnaissance, the primary task is to detect thermal targets against a dark, complex background. Therefore, an effective metric should measure the ``relative saliency'' of the target region with respect to its local neighborhood, rather than the global pixel variance. We formalize this principle into WLC.
	
	\subsection{Formulation of WLC}
\begin{figure}[htbp]
		\centering
		\includegraphics[width=0.85\linewidth]{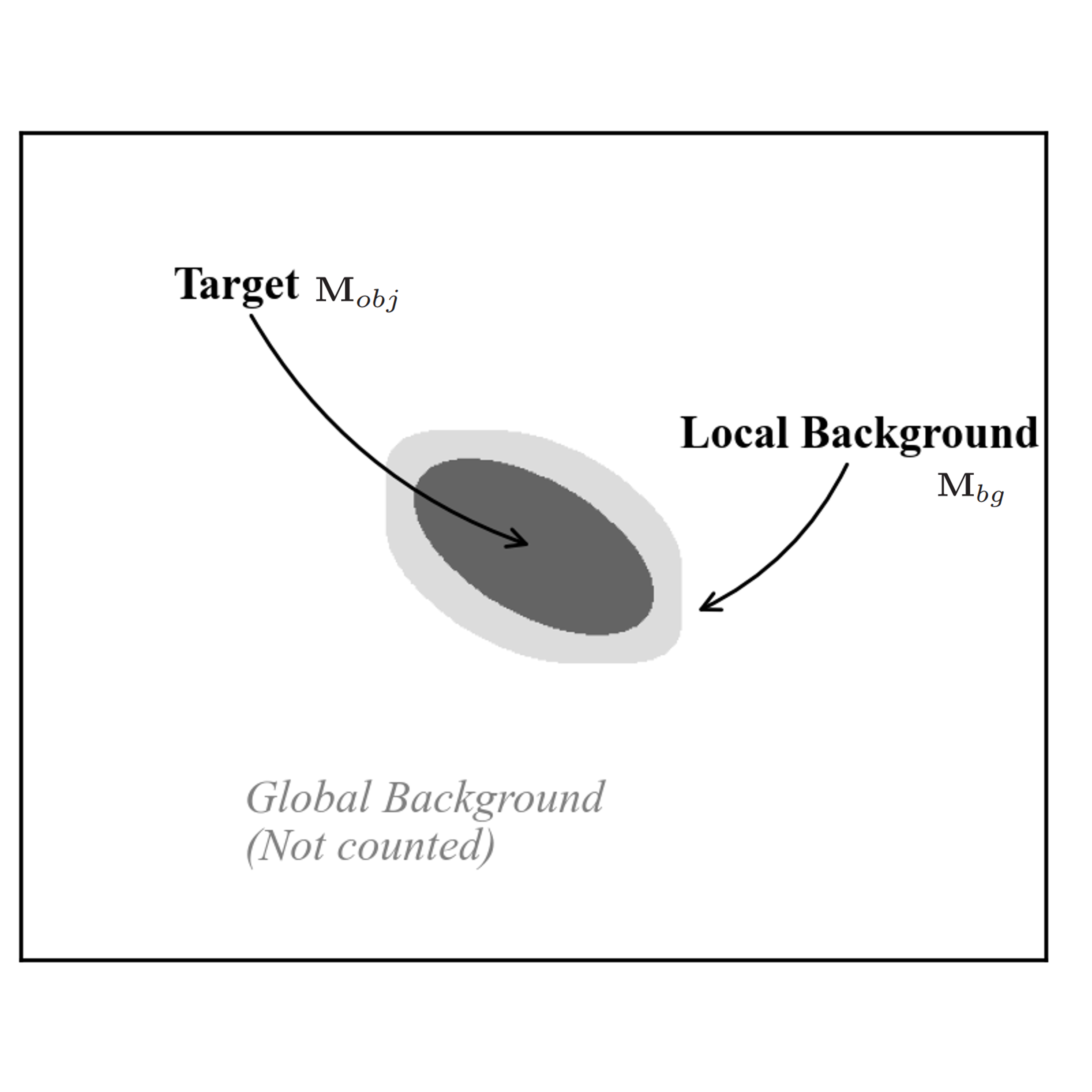}
		\caption{Schematic diagram of WLC calculation mechanism. The binary mask $\mathbf{M}_{obj}$ (dark gray) denotes the salient target extracted from the infrared prior. The annular region $\mathbf{M}_{bg}$ (light gray), obtained via morphological dilation, represents the local background. WLC measures the relative contrast between these two regions, explicitly excluding interference from the distant global background (white area).}
		\label{fig:TBC}
	\end{figure}

	The calculation of WLC involves three sequential stages: Salient Target Extraction, Local Background Modeling, and Contrast Computation. A schematic of WLC index calculation principle is shown in Fig. \ref{fig:TBC}.
	
	\textbf{Step 1: Target Mask Generation}
	
	Infrared images $\mathbf{I}_{ir}$ are the most reliable source for locating targets (e.g., vehicles, personnel). Since ground truth annotations are unavailable in unsupervised evaluation, this work utilizes the infrared image $\mathbf{I}_{ir}$ as a semantic prior. It utilizes infrared pixel intensity to generate a binary target mask $\mathbf{M}_{obj}$:
	\begin{equation}
		\mathbf{M}_{obj}(x,y) = 
		\begin{cases} 
			1, & \text{if } \mathbf{I}_{ir}(x,y) \ge T_{perc} \\
			0, & \text{otherwise} 
		\end{cases}
	\end{equation}
	where $T_{perc}$ is the $k$ percentile of the infrared pixel intensity (set to $k=98\%$ in this work, more details can be found in  \ref{sec:validation}). This ensures only the brightest thermal regions are selected.
	
	\textbf{Step 2: Local Background Extraction}
	
	Defining the ``immediate background'' is critical. Global background includes irrelevant clutter. By applying a morphological dilation operation $\mathfrak{D}$ to the target mask $\mathbf{M}_{obj}$ and subtracting the original target region, we obtain an annular background mask $\mathbf{M}_{bg}$ surrounding the target:
	\begin{equation}
		\mathbf{M}_{bg} = \mathfrak{D}(\mathbf{M}_{obj}, \mathcal{K}) - \mathbf{M}_{obj}
	\end{equation}
	where $\mathcal{K}$ is a $5 \times 5$ structural element. This definition avoids interference from distant background clutter, ensuring that contrast calculation occurs strictly between the target and its immediate surroundings.

	\textbf{Step 3: Contrast Calculation}
	
	Based on the generated masks, we calculate the average gray values of the fused image $\mathbf{F}$ in the target and background regions, respectively:
	\begin{equation}
		\mu_{obj} = \frac{\sum \mathbf{F} \cdot \mathbf{M}_{obj}}{\sum \mathbf{M}_{obj}}, \quad 
		\mu_{bg} = \frac{\sum \mathbf{F} \cdot \mathbf{M}_{bg}}{\sum \mathbf{M}_{bg}}
	\end{equation}
	Finally, WLC is defined as the relative luminance gain:
	\begin{equation}
		\mathrm{WLC} = \frac{\mu_{obj} - \mu_{bg}}{\mu_{bg} + \epsilon}
		\label{eg:tbc}
	\end{equation}
	where $\epsilon$ is a minimal constant to prevent division by zero. In addition, WLC describes the relative luminance gain.
	
	\subsection{Theoretical Properties of WLC}
%
%
WLC satisfies several desirable mathematical properties that ensure its reliability. Let $\mathbf{F}$ denote the fused image, with $\mu_{obj}$ and $\mu_{bg}$ representing the local mean intensities of the target and background regions, respectively. It should be noted that the following derivation uses (\ref{eg:tbc}), we require $\epsilon \to 0$.

\begin{property}[Noise Penalization]
	Under additive background noise, assuming the target intensity is preserved, the metric monotonically decreases: $\mathrm{WLC}(\tilde{\mathbf{F}}) < \mathrm{WLC}(\mathbf{F})$.
\end{property}

\begin{IEEEproof}
	Due to the non-negativity constraint of pixel intensities, zero-mean noise introduces a positive shift to the dark background mean, yielding $\tilde{\mu}_{bg} = \mu_{bg} + \delta$ ($\delta > 0$). Taking the partial derivative of WLC with respect to $\mu_{bg}$ yields:
	\begin{equation}
		\label{partial}
		\frac{\partial \mathrm{WLC}}{\partial \mu_{bg}} = \frac{-(\mu_{obj} + \epsilon)}{(\mu_{bg} + \epsilon)^2} < 0
	\end{equation}
	Since $\mu_{obj} \ge 0$ and $\epsilon > 0$, (\ref{partial}) is strictly negative. This proves that WLC actively penalizes noise.
\end{IEEEproof}

\begin{property}[Target Loss]
	If thermal saliency is entirely lost ($\mu_{obj} \to \mu_{bg}$), the metric converges to zero.
\end{property}

\begin{IEEEproof}
	If a fusion algorithm fails to preserve the target ($\mu_{obj} \to \mu_{bg}$), by direct substitution:
	\begin{equation}
		\label{target_loss}
		\lim_{\mu_{obj} \to \mu_{bg}} \mathrm{WLC} = \lim_{\mu_{obj} \to \mu_{bg}} \frac{\mu_{obj} - \mu_{bg}}{\mu_{bg} + \epsilon} = 0
	\end{equation}
	This property explicitly flags semantic loss, avoiding the false scores common in gradient-based metrics.
\end{IEEEproof}

\begin{property}[Scale Invariance]
	For any linear global illumination shift $\alpha > 0$, the metric remains invariant: $\mathrm{WLC}(\alpha \mathbf{F}) \approx \mathrm{WLC}(\mathbf{F})$.
\end{property}

\begin{IEEEproof}
	For the scaled image $\alpha \mathbf{F}$, region intensities scale proportionally.  By direct substitution
	\begin{equation}
		\mathrm{WLC}(\alpha \mathbf{F}) = \frac{\alpha \mu_{obj} - \alpha \mu_{bg}}{\alpha \mu_{bg} + \epsilon} \approx \frac{\mu_{obj} - \mu_{bg}}{\mu_{bg}}
	\end{equation}
	This formulation elegantly decouples absolute illumination variations, e.g., day-to-night transitions.
\end{IEEEproof}
	
	\section{Metric Validation}
	\label{sec:validation}
	Validating a metric without Ground Truth is challenging. Direct comparison on real algorithms risks circular reasoning. Therefore, this paper designs a controlled experiment by using synthetic degradation. The rationality of this experimental design is primarily reflected in the following two aspects. 
	
	First, the output of real fusion models typically mixes various distortions such as contrast loss, edge blurring, and residual noise. Second, by using synthetically generated cases, we can apply the control variable method to independently simulate single variables like ``target loss'', ``background noise'', and ``low contrast'', thereby precisely analyzing the metric's sensitivity to specific degradation types. By constructing clear ``Ideal Fusion'' and ``Severe Lost'' samples, we establish a scientific benchmark to satisfy the monotonicity assumption. That is, when image quality degrades significantly, the metric value should exhibit monotonic decay.
	
	\subsection{Experimental Setup}
	We select typical low-altitude night scenes from the DroneVehicle dataset \cite{sun2020drone} as a baseline to construct four representative cases: Ideal Fusion, Target Lost, Noisy, and Low Contrast. These four cases are synthetically generated as follows: 
	\begin{itemize}
		\item \textbf{Ideal:} This case integrates the raw infrared target directly into a clean, noise-free visible background, representing the theoretical upper bound of fusion quality.
		\item \textbf{Lost:} This case retains the visible background but significantly attenuates the infrared target intensity, simulating the failure to preserve thermal saliency.
		\item \textbf{Noisy:} Gaussian noise is injected into the Ideal case to simulate high-frequency sensor artifacts common in low-light conditions.
		\item \textbf{Low Contrast:} Both the target and background intensities are artificially brightened, reducing the relative saliency and simulating overexposure.
	\end{itemize}	
	
\begin{figure*}[!htbp]
		\centering
		\includegraphics[width=1.0\linewidth]{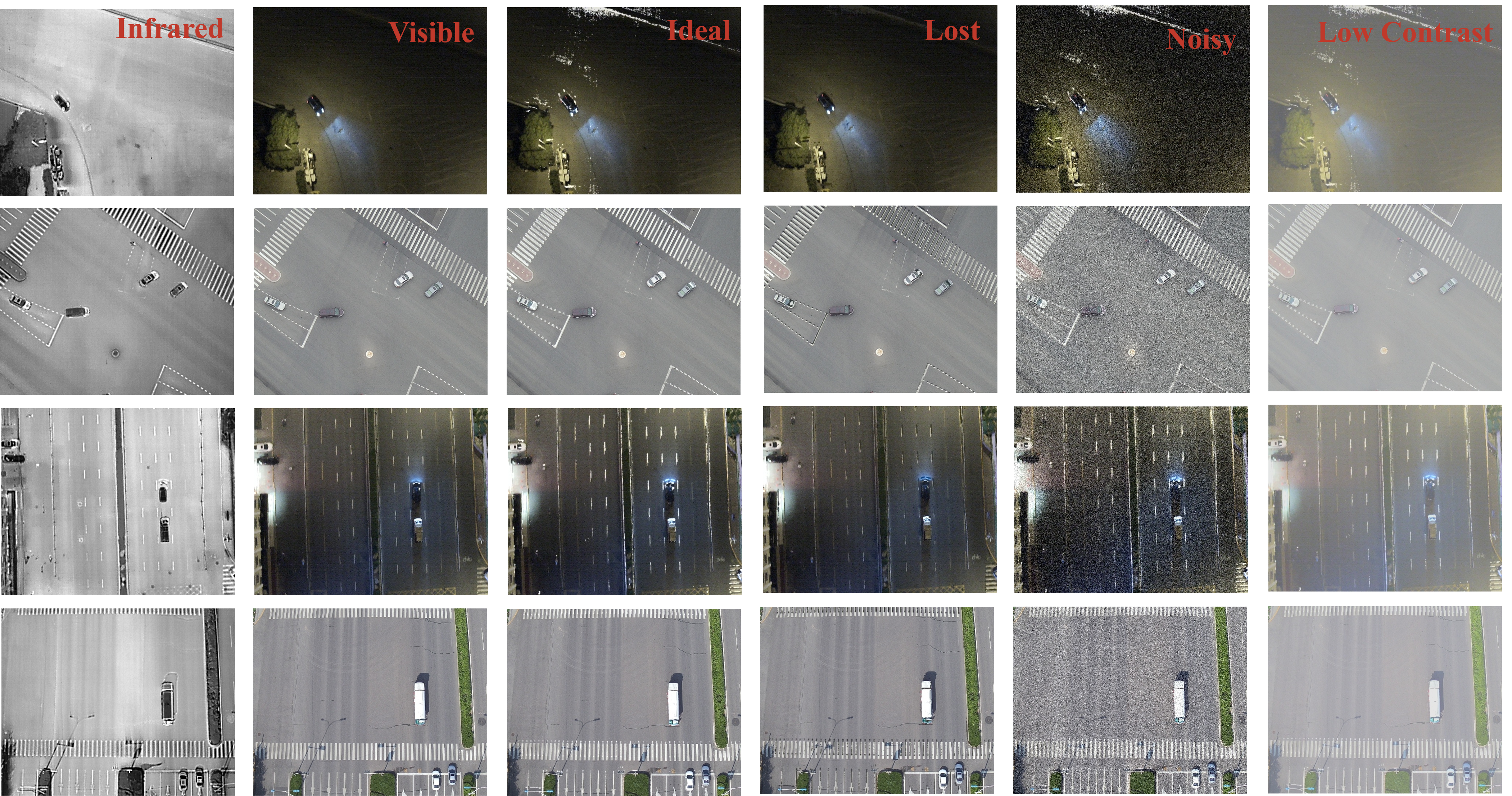}
		\caption{Visualization of different image types for samples (from top to bottom are $00942$, $02994$, $03813$, and $04834$) from the DroneVehicle dataset. Columns represent infrared, visible, ideal, lost, noisy, and low contrast images respectively.}
		\label{fig:dronevehicle_samples}
	\end{figure*}

	To ensure generalization, this paper conducts tests on the entire DroneVehicle dataset. We choose only four samples to illustrate our concepts , as shown in Fig. \ref{fig:dronevehicle_samples}. Moreover, we use data processing methods such as box plots to present the indicator results. We also verify the formula (\ref{eg:tbc}) through noise degradation experiments. Furthermore, this paper discusses the selection of parameter values and time complexity.
	
	All experiments are conduct on a standard computing platform equipped with an Intel Core i5-14500 CPU (2.60 GHz), ensuring the reproducibility of the efficiency benchmarks.

	\subsection{Result Analysis}
	To intuitively demonstrate the response of different evaluation paradigms to image degradation, we first conduct a fine-grained analysis on four representative samples, including low-light night scenes (Samples $00942$ and $02994$) and complex daytime scenarios (Samples $03813$ and $04834$). As detailed in Table \ref{tab:combined_samples}, a cross-comparison reveals a catastrophic failure mode inherent in traditional metric families. 
	
	Specifically, gradient-based metrics (AG and SF) exhibit a pathological sensitivity to high-frequency interference. For instance, in the `Noisy' case of Sample $04834$, the scores of AG and SF surge to $0.524$ and $0.311$, approximately $3$ to $4$ times higher than their scores in the `Ideal' case. This empirically confirms that gradient operators are mathematically incapable of distinguishing between the structural edges of the scene and the random fluctuations of noise. Similarly, statistics-based metrics (SD and EN) consistently assign their highest scores to the `Noisy' case (e.g., EN rises to $7.534$), indicating that these metrics misinterpret noise-induced histogram spreading as valid information gain. 
	In stark contrast, the proposed WLC metric demonstrates remarkable robustness. While random noise fluctuations may cause slight oscillations due to the stochastic brightening of pixels, WLC effectively avoids the ``score explosion'' observed in AG. More importantly, in the `Lost' case where thermal signatures are attenuated, WLC responds with negative or near-zero values (e.g., $-0.383$ for Sample $04834$), effectively flagging the semantic loss of targets—a critical capability completely absent in EN and SD.

	\begin{table}[t]
		\centering
		\caption{Quantitative Comparison Across Four Diverse Samples}
		\label{tab:combined_samples}

		\renewcommand{\arraystretch}{1.1} 
		\small
		\begin{threeparttable}
		\begin{tabular}{ccccccc}
			\toprule
			\multirow{2}{*}{\textbf{Sample}} & \multirow{2}{*}{\textbf{Case}} & \textbf{WLC} & \multirow{2}{*}{\textbf{AG}} & \multirow{2}{*}{\textbf{SD}} & \multirow{2}{*}{\textbf{SF}} & \multirow{2}{*}{\textbf{EN}} \\
			&  & \scriptsize{(Ours)} &  &  &  &  \\
			\midrule
			
			\multirow{4}{*}{$00942$} 
			& Ideal & \textbf{$2.185$} & $0.122$ & $0.130$ & $0.057$ & $6.619$ \\
			& Lost  & $-0.081$ & $0.089$ & $0.108$ & $0.037$ & $6.478$ \\
			& Noisy & \textcolor{red}{$2.191$} & \textcolor{red}{\textbf{$0.491$}} & \textcolor{red}{\textbf{$0.198$}} & \textcolor{red}{\textbf{$0.305$}} & \textcolor{red}{\textbf{$6.621$}} \\
			& LowCon & $0.456$ & $0.061$ & $0.065$ & $0.029$ & $5.620$ \\
			\cmidrule{1-7}
			
			\multirow{4}{*}{$02994$} 
			& Ideal & \textbf{$0.507$} & $0.118$ & $0.090$ & $0.063$ & $5.674$ \\
			& Lost  & $-0.400$ & $0.121$ & $0.077$ & $0.067$ & $5.659$ \\
			& Noisy & \textcolor{red}{$0.508$} & \textcolor{red}{\textbf{$0.497$}} & \textcolor{red}{\textbf{$0.175$}} & \textcolor{red}{\textbf{$0.307$}} & \textcolor{red}{\textbf{$7.426$}} \\
			& LowCon & $0.217$ & $0.059$ & $0.045$ & $0.032$ & $4.678$ \\
			\cmidrule{1-7}
			
			\multirow{4}{*}{$03813$} 
			& Ideal & \textcolor{red}{\textbf{$1.461$}} & $0.167$ & $0.156$ & $0.074$ & $6.906$ \\
			& Lost  & $-0.184$ & $0.134$ & $0.133$ & $0.056$ & $6.767$ \\
			& Noisy & $1.453$ & \textcolor{red}{\textbf{$0.516$}} & \textcolor{red}{\textbf{$0.216$}} & \textcolor{red}{\textbf{$0.308$}} & \textcolor{red}{\textbf{$7.320$}} \\
			& LowCon & $0.422$ & $0.084$ & $0.078$ & $0.037$ & $5.907$ \\
			\cmidrule{1-7}
			
			\multirow{4}{*}{$04834$} 
			& Ideal & \textbf{$0.470$} & $0.176$ & $0.125$ & $0.083$ & $6.258$ \\
			& Lost  & $-0.383$ & $0.180$ & $0.116$ & $0.085$ & $6.234$ \\
			& Noisy & \textcolor{red}{$0.474$} & \textcolor{red}{\textbf{$0.524$}} & \textcolor{red}{\textbf{$0.195$}} & \textcolor{red}{\textbf{$0.311$}} & \textcolor{red}{\textbf{$7.534$}} \\
			& LowCon & $0.202$ & $0.088$ & $0.062$ & $0.041$ & $5.261$ \\
			\bottomrule
		\end{tabular}
		\begin{tablenotes} 
			\footnotesize
			\item[1] Red highlighting is the highest value in the column, the same applies throughout this paper.
		\end{tablenotes}
		\end{threeparttable}
	\end{table}

	\begin{table}[ht] 
		\centering
		\caption{Statistical Performance Analysis ($N=300$).}
		\label{tab:metric_performance}
		
		\renewcommand{\arraystretch}{1.3} 
		\setlength{\tabcolsep}{1.5pt}      
		\small
		
		\resizebox{\columnwidth}{!}{
			\begin{tabular}{c ccccc}
				\toprule
				\textbf{Case} & \shortstack{\textbf{WLC}\\ \scriptsize{(Ours)}} & \textbf{AG} & \textbf{SF} & \textbf{EN} & \textbf{SD} \\
				\midrule
				Ideal & \textcolor{red}{\textbf{$3.04 \pm 3.95$}} & $0.16 \pm 0.06$ & $0.07 \pm 0.02$ & $6.44 \pm 1.11$ & $0.16 \pm 0.04$ \\
				Lost & $0.43 \pm 1.70$ & $0.14 \pm 0.07$ & $0.06 \pm 0.03$ & $6.34 \pm 1.14$ & $0.13 \pm 0.06$ \\
				Noisy & $2.02 \pm 1.87$ & \textcolor{red}{\textbf{$0.45 \pm 0.08$}} & \textcolor{red}{\textbf{$0.27 \pm 0.04$}} & \textcolor{red}{\textbf{$6.69 \pm 1.08$}} & \textcolor{red}{\textbf{$0.20 \pm 0.04$}} \\
				LowCon & $0.43 \pm 0.19$ & $0.08 \pm 0.03$ & $0.04 \pm 0.01$ & $5.44 \pm 1.11$ & $0.08 \pm 0.02$ \\
				\bottomrule
			\end{tabular}
		}
	\end{table}
	
\begin{figure*}[!h]
		\centering
		\includegraphics[width=0.85\textwidth]{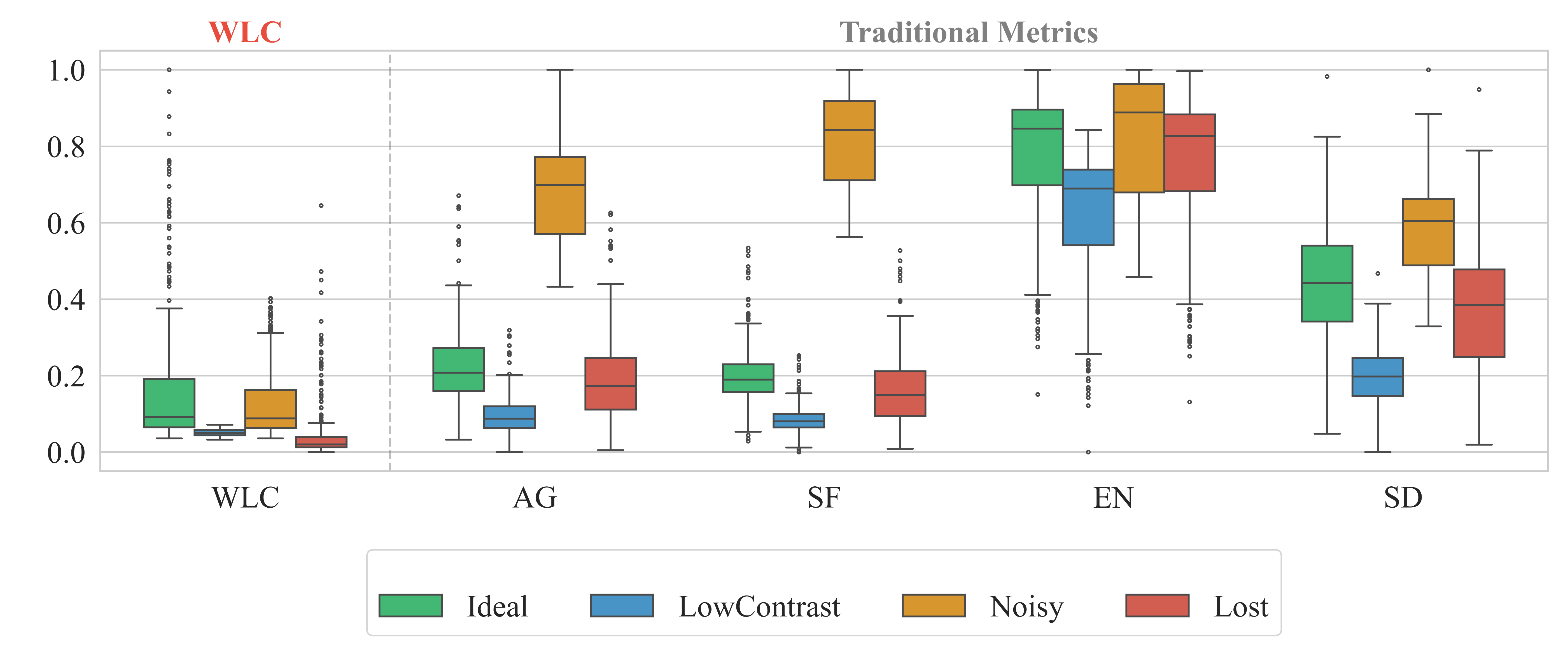}
		\caption{Systematic comparison of metric responses to synthetic degradation ($N=300$). The boxplots visualize the statistical distribution of scores for the proposed WLC and four traditional metrics across diverse cases.
			(Left) WLC correctly assigns significantly higher scores to the `Ideal' case (green) compared to the `Noisy' case (yellow), demonstrating its valid discriminative power.
			(Right) Traditional metrics (AG, SF, EN, and SD) exhibit a systematic failure, consistently rating the `Noisy' case higher than or equal to the `Ideal' case, confirming their susceptibility to ``Noise Trap''.}
		\label{fig_box}
	\end{figure*}

	To rule out aleatory uncertainty associated with single-sample evaluations, we extended the validation to a macroscopic statistical analysis of $300$ randomly selected samples from the DroneVehicle dataset. The aggregated results, summarized in Table \ref{tab:metric_performance} and visualized in Fig. \ref{fig_box}, provide decisive evidence of the metric's validity. 
	
	WLC emerges as the sole metric that correctly ranks the `Ideal' case ($\bar{x}=3.04$, $\bar{x}$ is statistical mean.) significantly higher than the `Noisy' case ($\bar{x}=2.02$). This statistical trend confirms that, at the dataset level, WLC effectively penalizes background noise through its local contrast formulation. Conversely, AG, SF, SD, and EN all incorrectly identify the `Noisy' case as the best-performing category (highlighted in red), corroborating that ``Noise Trap'' is a systematic defect rather than an isolated anomaly.

	Furthermore, the analysis of variance offers profound insights into the physical interpretability of the metrics. A crucial observation from Table \ref{tab:metric_performance} is that WLC exhibits a relatively high variance in the `Ideal' case ($\sigma=3.95$, $\sigma$ is statistical standard deviation). From a theoretical perspective, this high variance in WLC should be interpreted as ``Semantic Discriminability'' rather than instability. The DroneVehicle dataset inherently contains targets with vast differences in thermal intensity, ranging from active engines of trucks to cold little vehicles. WLC physically measures this relative contrast, thus faithfully reflecting the real-world diversity of thermal signatures. In contrast, traditional metrics show extremely low variance. The low variance suggests a ``Semantic Numbness''. These metrics treat different semantic targets and random noise indiscriminately as gradients or entropies. This is fatal for low-altitude tasks.

	\subsection{Discussion}
	This section supplements the previous experiments with comprehensive discussions on noise degradation, sensitivity analysis, and time complexity.
	
	\subsubsection{Noise Degradation}
	Previous studies on image denoising have demonstrated that handling mixed or unknown noise is a critical requirement for image quality preservation \cite{ref50,ref22}. As image quality objectively degrades, the metric score should decrease monotonically. To verify this, this paper conducts noise degradation experiments on four representative samples (Sample $00942$, $02994$, $03813$, and $04834$) by adding Gaussian noise with increasing standard deviation $\sigma \in [0,0.2]$.
	\begin{figure}[!h]  
		\centering
		\setlength{\tabcolsep}{3pt}  
		\renewcommand{\arraystretch}{0.8}  
		\renewcommand{\thesubfigure}{}
		\captionsetup[subfloat]{labelformat=empty} 
		
		\subfloat[]{
			\includegraphics[width=0.45\linewidth]{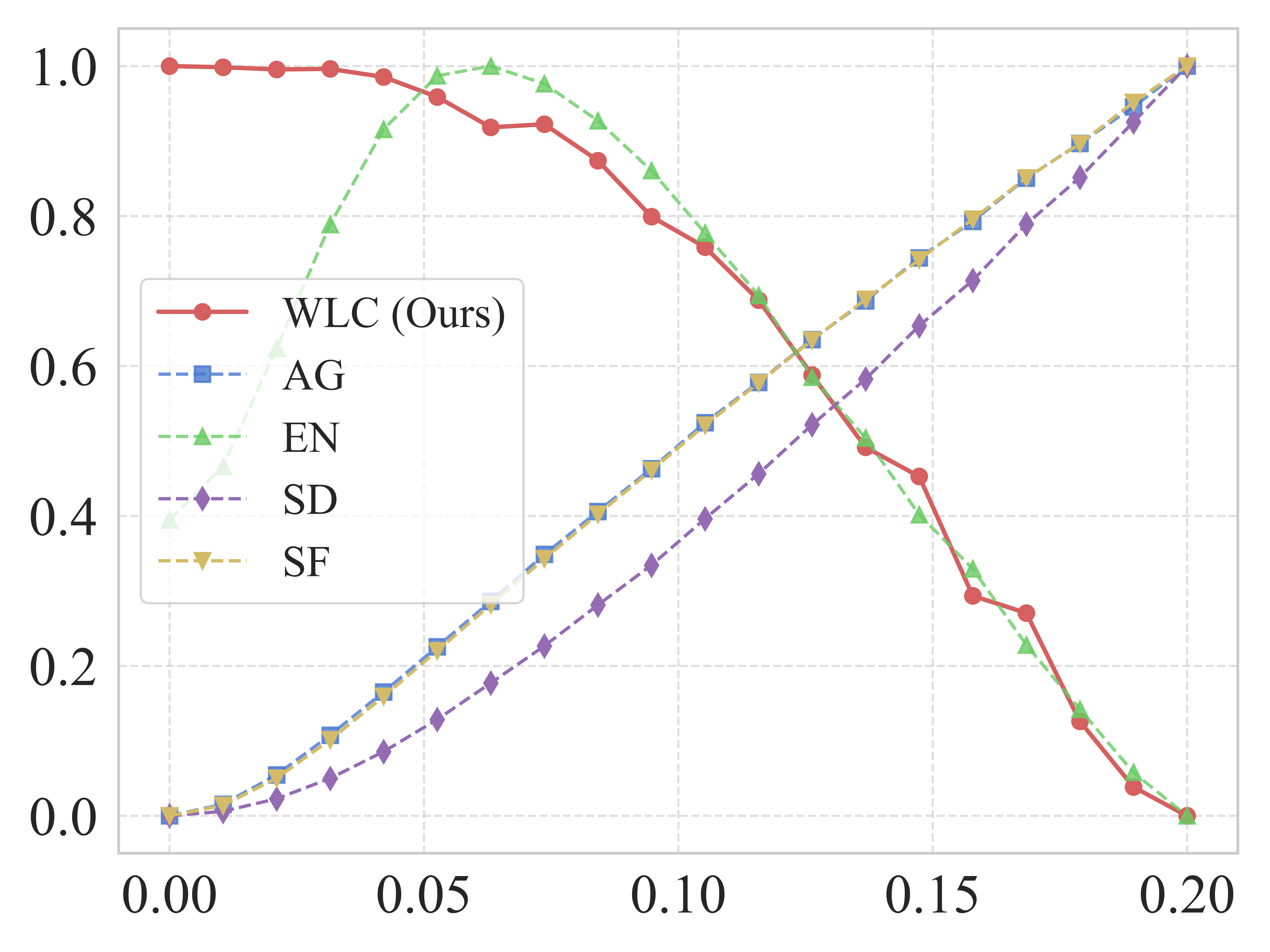}
		}
		\hfil  
		\subfloat[]{
			\includegraphics[width=0.45\linewidth]{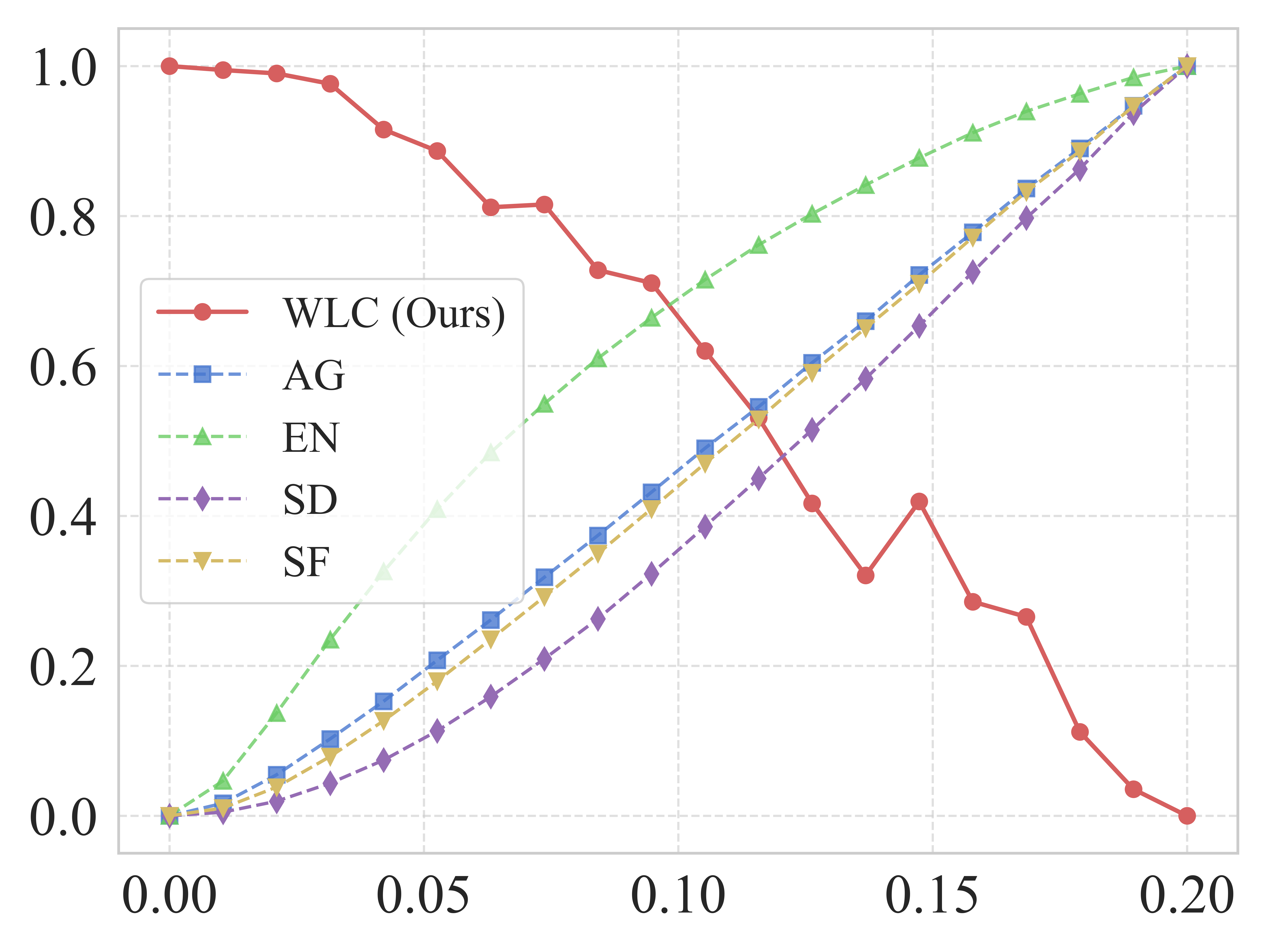}
		}
		
		\vspace{0.1em}  
		
		\subfloat[]{
			\includegraphics[width=0.45\linewidth]{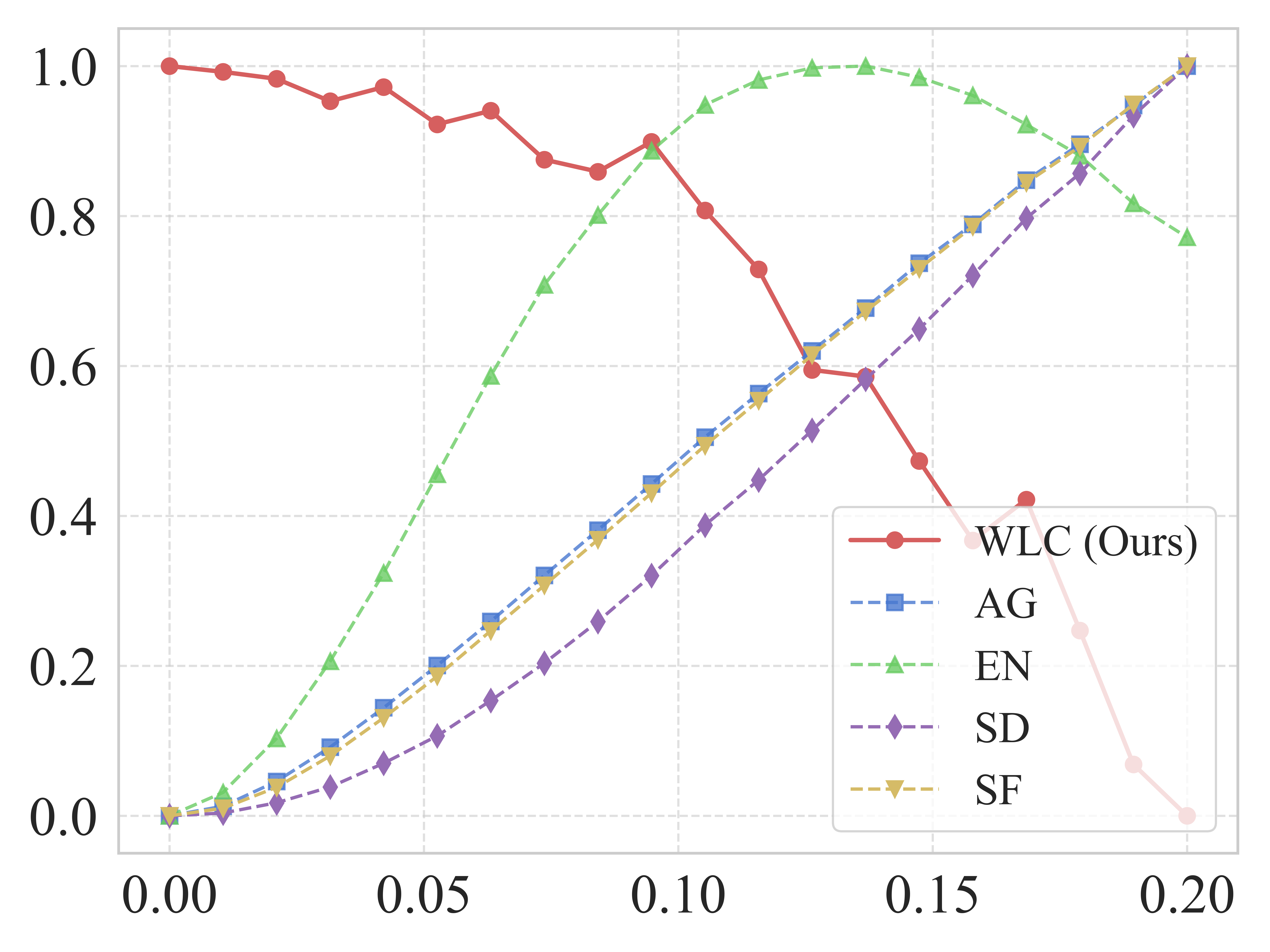}
		}
		\hfil
		\subfloat[]{
			\includegraphics[width=0.45\linewidth]{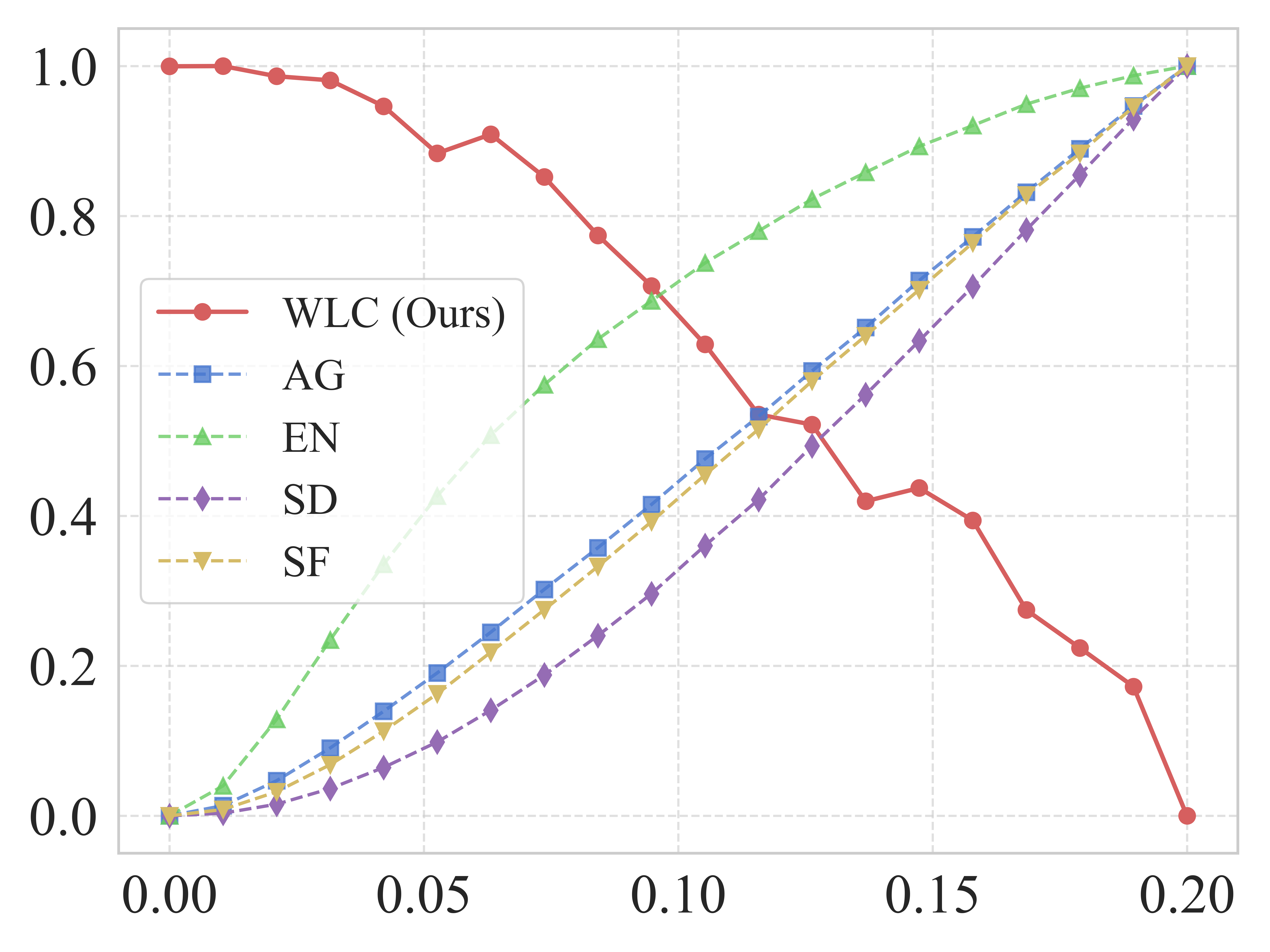}
		}
		
		\caption{Single-run noise degradation response curves for four diverse samples. Traditional metrics (AG, EN, SD, and SF) monotonically increase, indicating a failure to penalize noise. WLC's curve shows an overall downward trend but exhibits localized fluctuations due to the stochastic nature of noise within small local target regions.}
		\label{noise_bad}
		
	\end{figure}
	
	\begin{figure}[!h]
		\centering
		\setlength{\tabcolsep}{3pt}
		\renewcommand{\arraystretch}{0.8}
		\renewcommand{\thesubfigure}{}  
		\captionsetup[subfloat]{labelformat=empty} 
		
		\subfloat[]{%
			\includegraphics[width=0.45\linewidth]{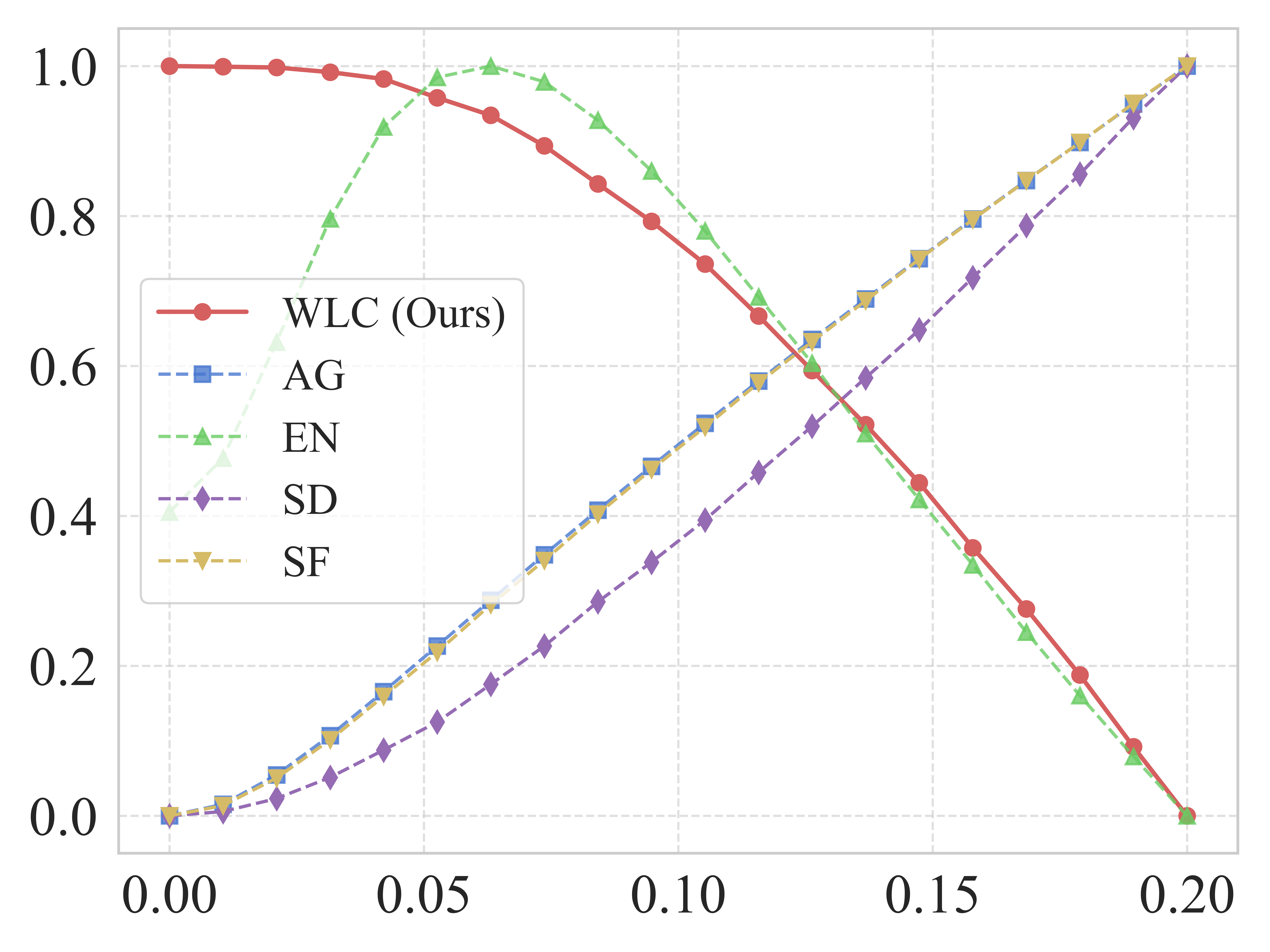}%
		}
		\hfil
		\subfloat[]{%
			\includegraphics[width=0.45\linewidth]{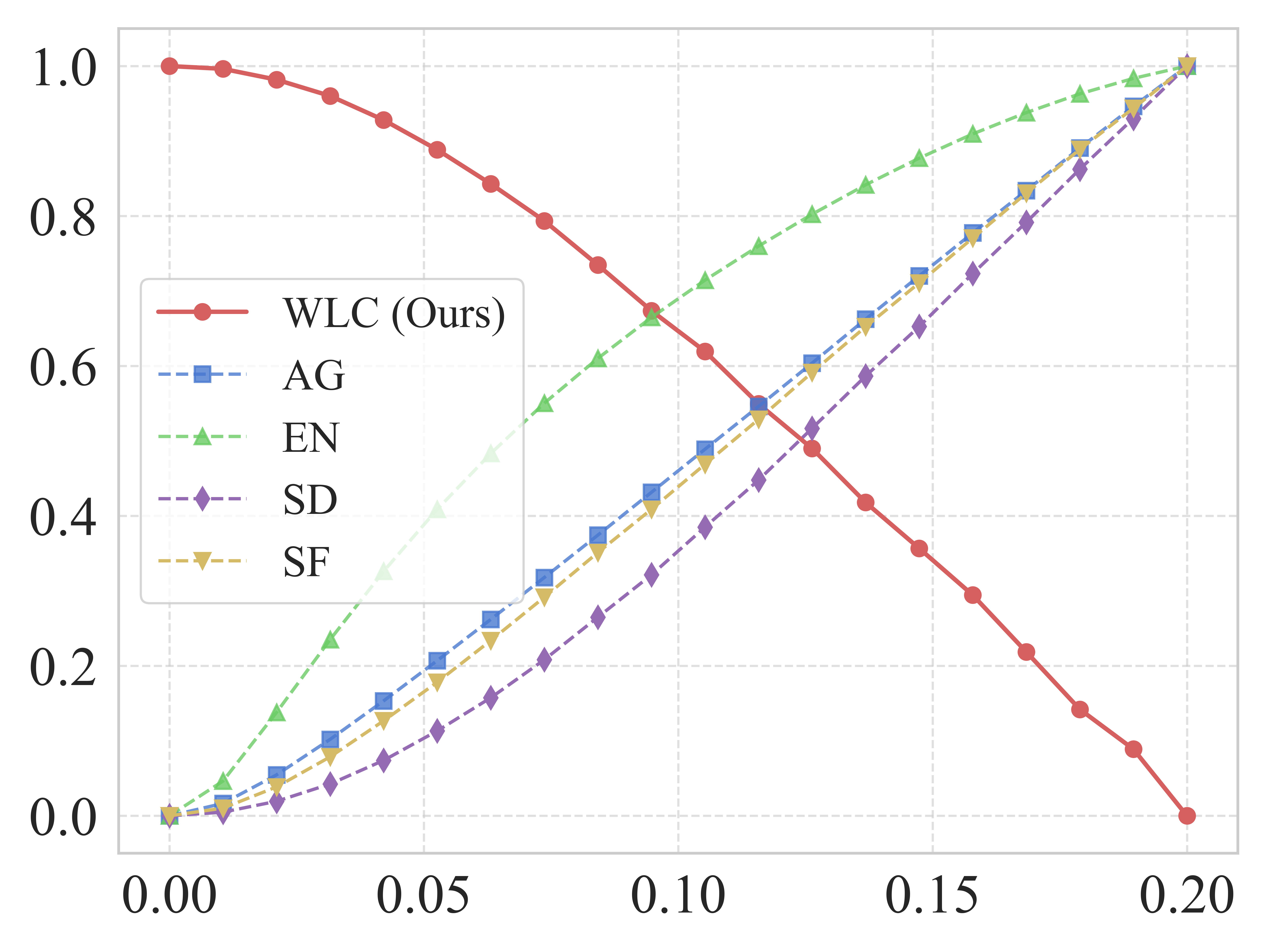}%
		}
		
		\vspace{0.05em}
		
		\subfloat[]{
			\includegraphics[width=0.45\linewidth]{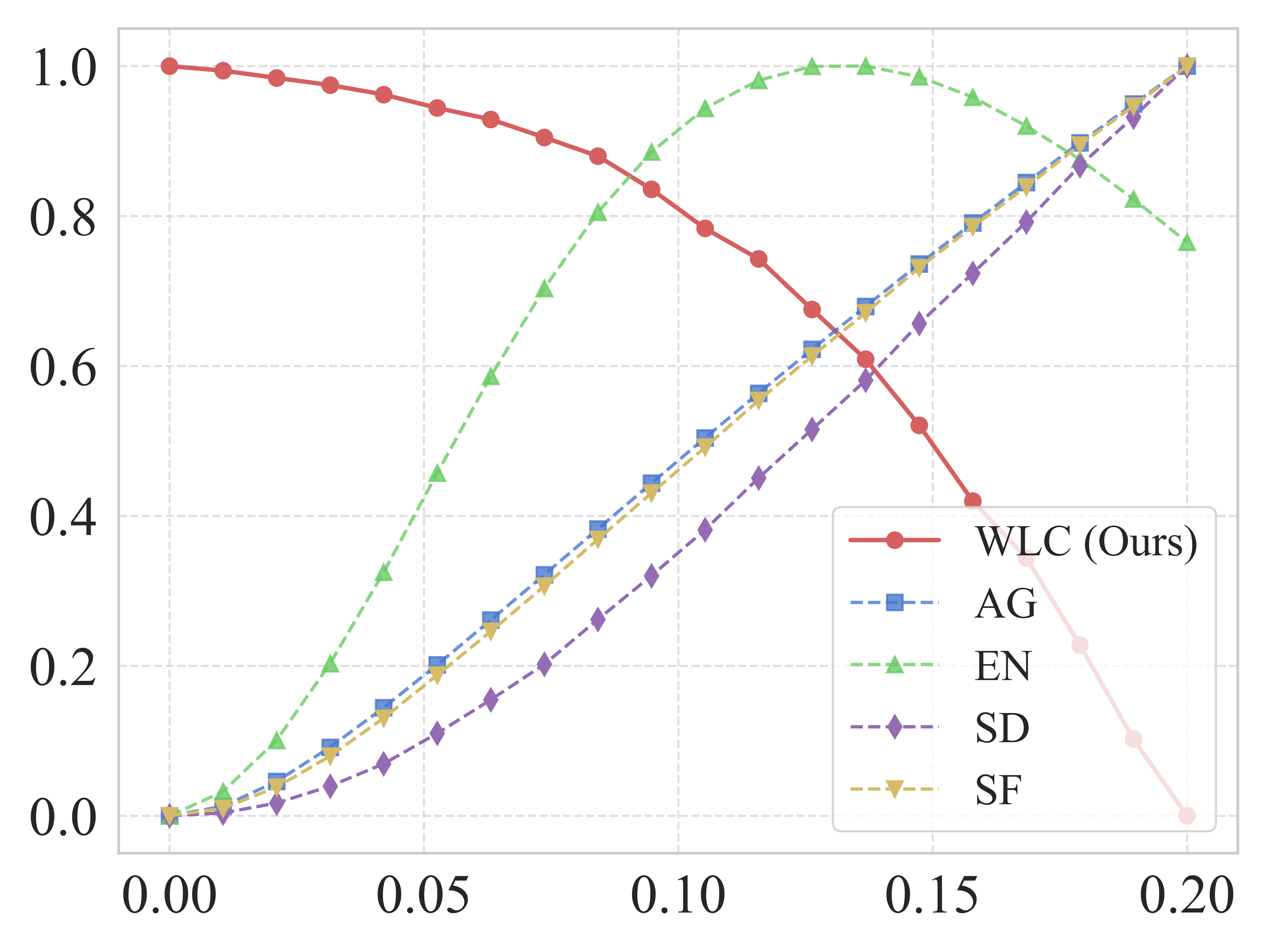}%
		}
		\hfil
		\subfloat[]{
			\includegraphics[width=0.45\linewidth]{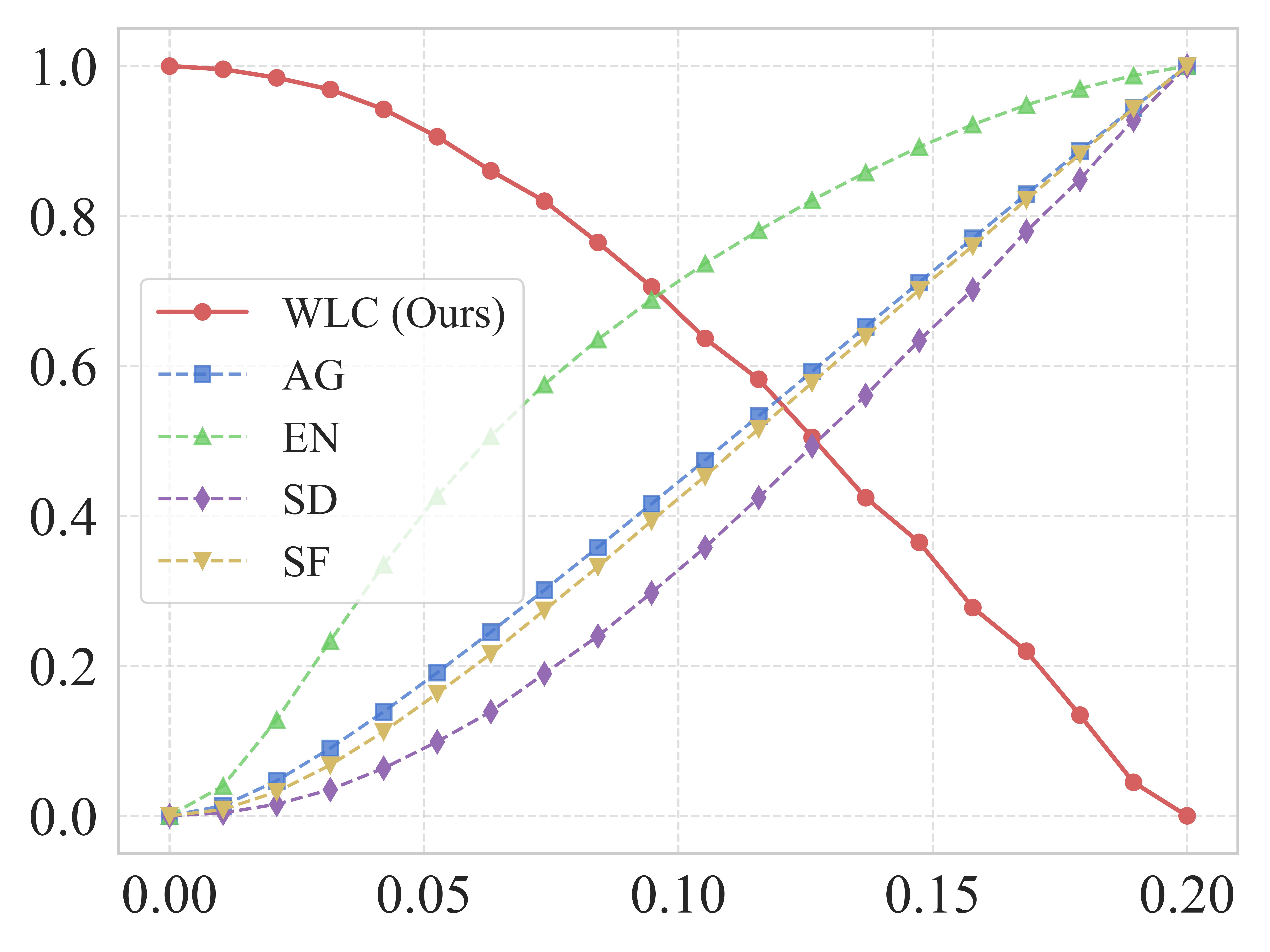}%
		}
		
		\caption{Monte Carlo averaged noise degradation curves ($N=50$). By averaging out aleatory uncertainty, the underlying statistical trends are revealed. WLC demonstrates a smooth, monotonic decline, confirming its robustness, whereas traditional metrics consistently fall into ``Noise Trap''.}
		\label{noise_good}
	\end{figure}
	
	Figs. \ref{noise_bad} and \ref{noise_good} demonstrate that traditional metrics fall into ``Noise Trap''. The curves for AG, SF, and SD (represented by blue, yellow, and purple dashed lines) exhibit a strictly monotonic upward trend across all samples. The EN curve shows a rapid logarithmic increase. 
	As mentioned in Section \ref{sec:method}, AG, SF, and SD are linearly related to noise. AG and SF are almost equivalent. Moreover, the AG score scales proportionally with $\sqrt{\pi} \sigma$. SD, however, shows a purely linear relationship. Thus, the two lines of AG and SF almost coincide and are higher than SD. 
	
	In sharp contrast, WLC demonstrates a consistent monotonic decline across all samples. As noise levels rise, the local background intensity $\mu_{bg}$ increases while the target saliency becomes corrupted, leading to a natural penalty in WLC score. This trend aligns perfectly with human visual perception, confirming WLC's ability to distinguish signal from noise.
	
	It is worth noting the difference between the single-run results (Fig. \ref{noise_bad}) and the Monte Carlo averaged results (Fig. \ref{noise_good}). 
	In single-run experiments, WLC's curves exhibit localized stochastic oscillations (jaggedness), particularly at high noise levels. This phenomenon is not an instability but reflects the local nature of WLC. In the previous analysis, it was assumed that the noise follows a normal distribution. Therefore, global metrics average statistics over millions of pixels. According to the law of large numbers, the likelihood of errors becomes very small. However,  WLC calculates contrast based on small target neighborhoods which often $\leq 100$ pixels. Consequently, it is highly sensitive to the random polarity of noise. 
	
	To mitigate this aleatory uncertainty and reveal the underlying statistical trend, this paper uses the Monte Carlo mean with 50 repetitions. As shown in Fig. \ref{noise_good}, the averaged curves become smooth and strictly monotonic. This proves that while WLC is sensitive to local pixel variations, its statistical expectation is robust and theoretically sound.
	
	Another phenomenon worth noting is that a counter-intuitive ``inverted-U'' trend is observed in the EN curve , as shown in Fig. \ref{noise_good}. Initially, the metric score rises as expected, but it begins to decline after a certain noise threshold.
	This phenomenon is caused by the Pixel Saturation Effect subject to the limited dynamic range $[0, 255]$. 
	As noise intensity escalates, a significant portion of pixel values exceed the physical bounds. The mandatory clipping operation forces these outliers to accumulate at the boundaries ($0$ and $255$), transforming the pixel probability distribution from a spread-out Gaussian-like form to a polarized bimodal distribution. 
	Mathematically, this boundary accumulation reduces the statistical uncertainty of the histogram, paradoxically lowering the entropy score.
	This non-monotonicity is a critical flaw. It implies that EN suffers from ambiguity, as a single score could correspond to either a moderately detailed image or a heavily saturated noisy image, rendering it unreliable for optimization.

	\subsubsection{Sensitivity Analysis}
\begin{figure}[htbp]
		\centering
		\includegraphics[width=0.95\linewidth]{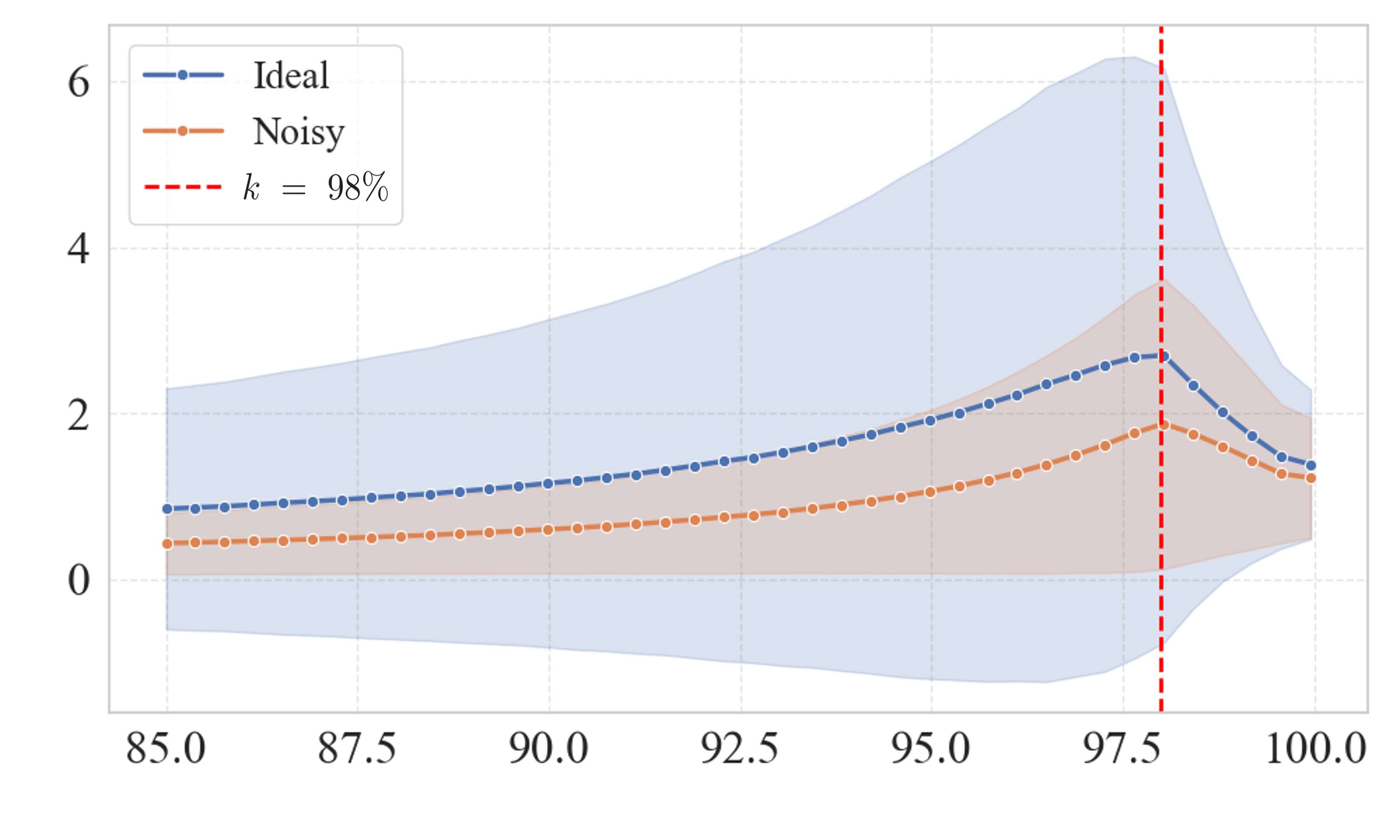}
		\caption{Threshold sensitivity analysis of WLC. The curves illustrate  WLC scores across varying percentile thresholds for `Ideal' and `Noisy' cases. The distinct inverted-U shape confirms the existence of an optimal operating point. The performance peaks at $k\approx 98\%$. Lower thresholds ($k \leq 95\%$) dilute contrast with background noise, while higher thresholds ($k > 99\%$) cause score collapse due to target loss.}
		\label{fig:sensitivity}
	\end{figure}

	WLC relies on a hyperparameter $k$ (set to $98\%$ by default) to determine the infrared saliency threshold. 
	To investigate the robustness of this choice, this paper analyzes the metric's response across a wide range of percentiles $k \in [85\%, 99.5\%]$.
	Fig. \ref{fig:sensitivity} illustrates WLC scores for both the `Ideal' and `Noisy' synthetic cases. This curve exhibits a distinct inverted U-shaped distribution.
	
	In the range of $k \in [85\%, 95\%]$, the scores rise gradually as the threshold effectively filters out low-intensity background clutter. In the range of $k \in [95\%, 99\%]$, the metric reaches a performance plateau, where WLC scores are maximized. This indicates a ``Robust Interval'' where the metric effectively captures the target's core thermal signature while excluding noise. Beyond $k > 99\%$, the scores drop sharply because the mask becomes too sparse to represent the integrity of the target.
	
	Therefore, while any value within $[95\%, 99\%]$ yields satisfactory results, this paper recommends $k=98\%$ as the default configuration as a priori knowledge. Finally, the DroneVehicle dataset inherently contains targets with a vast dynamic range of thermal intensities. As mentioned above, WLC faithfully records this diversity, resulting in a wide score distribution for clean images.
	
	To prevent evaluation bias and demonstrate broad applicability, this paper extends the sensitivity analysis to two additional benchmark datasets: TNO \cite{TNO} and M3FD \cite{M3FD}. This paper apply the same synthetic degradation protocol (Ideal vs. Noisy) across a wide range of thresholds $k \in [85\%, 100\%]$. 
	
	\begin{figure}[!t]
		\centering
		\subfloat[\footnotesize TNO Dataset]
		{
			\includegraphics[width=0.47\linewidth]{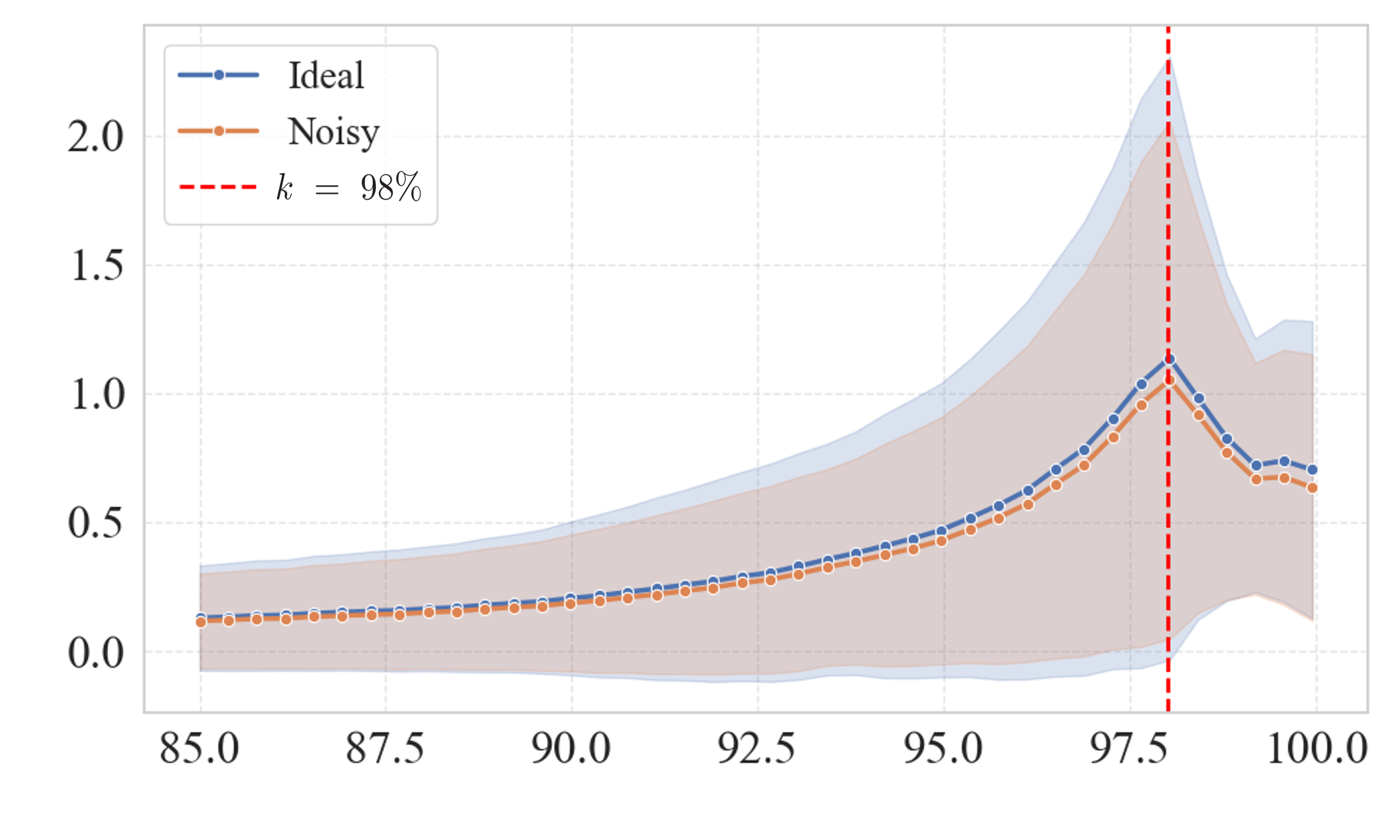} 
			\label{fig:sens_tno}
		}
		\hfil
		\subfloat[\footnotesize M3FD Dataset]
		{
			\includegraphics[width=0.47\linewidth]{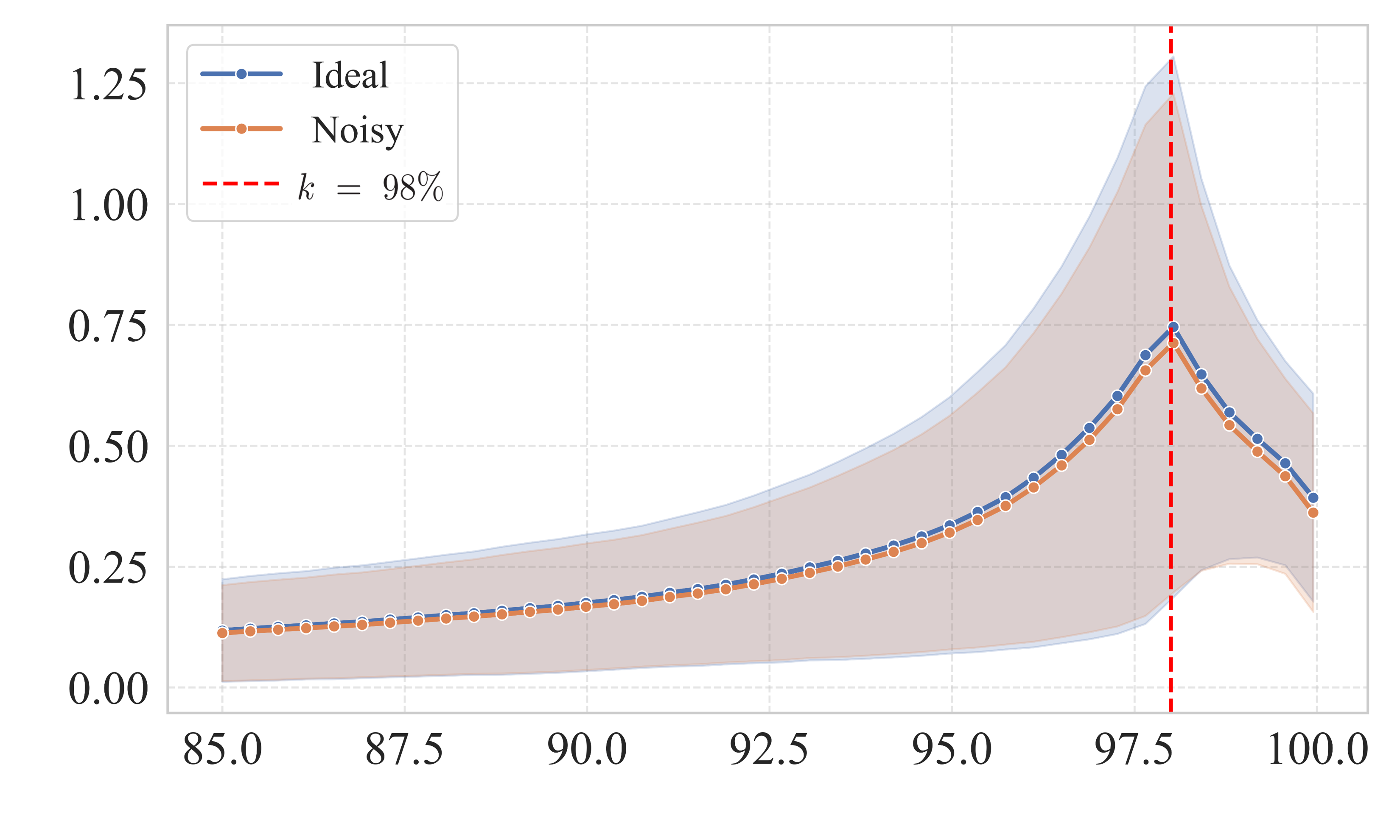} 
			\label{fig:sens_m3fd}
		}
		\caption{Cross-dataset parameter sensitivity analysis. Despite differences in sensor resolution and scene content, WLC scores on both (a) TNO and (b) M3FD datasets exhibit a consistent inverted-U trend, peaking at $k \approx 98\%$. This confirms that the proposed threshold is a prior parameter reflecting thermal saliency.}
		\label{fig:generalization}
	\end{figure}
	
	As visualized in Fig. \ref{fig:generalization}, despite the significant differences in scene content (e.g., soldiers in TNO vs. vehicles in M3FD), the sensitivity curves exhibit a remarkable consistency with the DroneVehicle results. These results perfectly align with the previous conclusions.
	
	It is noteworthy that the standard deviation (indicated by the shaded region) exhibits a sharp peak coincident with the mean score at $k=98\%$. In particular, in Fig. \ref{fig:generalization}, the blue and red areas almost overlap. TNO and M3FD are classic infrared datasets, and infrared cameras generally have a better signal-to-noise ratio than DroneVehicle. This means that the boundaries between the target and the context are very clear. Thus, WLC is sufficiently sensitive to the difference between highly salient targets (e.g., active vehicles) and subtle thermal signatures, thereby maximizing the statistical variance. Therefore, it appears ``sharp'' rather than ``rounded''.
	
	\subsubsection{Time Complexity}
	To validate the real-time capability of WLC for edge-device deployment, this paper benchmarks its execution time against traditional metrics across three resolutions ($256^2, 512^2, 1024^2$). Similarly, this paper selects the previously mentioned real-world samples and randomly crops them to the target size for experiments. To eliminate the randomness of experiments, this paper also randomly generates experimental data for comparison.
	\begin{figure}[htbp]
		\centering
		\includegraphics[width=1\linewidth]{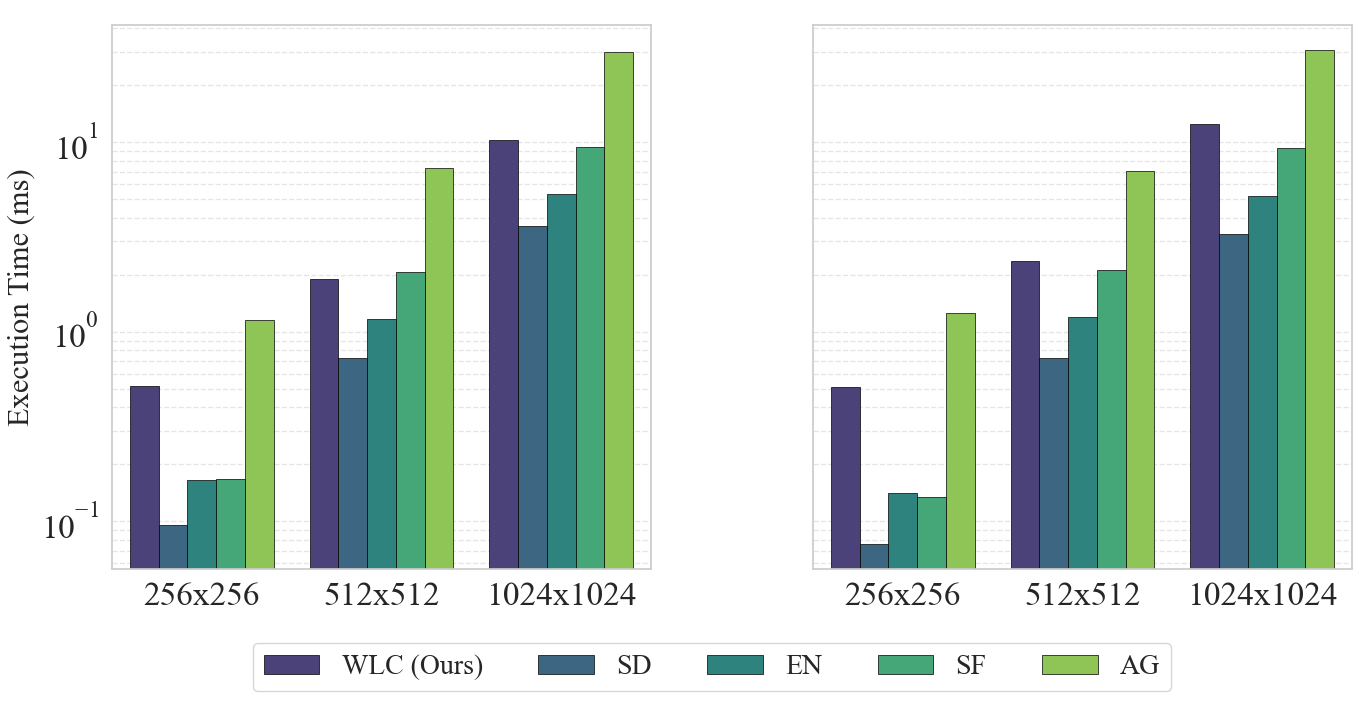}
		\caption{Dual-validation of computational complexity across different resolutions. The grouped bar charts (log scale) compare the average execution time of WLC against traditional metrics using both (Left) $300$ real-world samples and (Right) $100$ Monte Carlo simulation runs.}
		\label{fig:complex}
	\end{figure}

	\begin{table}[h]
		\centering
		\caption{Detailed Runtime Data (Unit: ms)}  
		\label{tab:runtime_combined}  
		
		\setlength{\tabcolsep}{3pt} 
		
		\renewcommand{\arraystretch}{1.2} 
		
		\resizebox{\linewidth}{!}{
			\begin{tabular}{lcccccc}  
				\toprule
				
				\textbf{Category} & \textbf{Resolution} & \textbf{AG} & \textbf{EN} & \textbf{SD} & \textbf{SF} & \shortstack{\textbf{WLC}\\ \scriptsize{(Ours)}} \\
				\midrule
				\multirow{3}{*}{Real-World} 
				& $256 \times 256$  & $1.154$ & $0.173$ & $0.094$ & $0.151$ & $0.515$ \\
				& $512 \times 512$  & $8.916$ & $1.333$ & $0.875$ & $2.396$ & $2.100$ \\
				& $1024 \times 1024$ & \textcolor{red}{\textbf{$70.546$}} & \textcolor{red}{\textbf{$11.962$}} & \textcolor{red}{\textbf{$8.371$}} & \textcolor{red}{\textbf{$21.461$}} & \textcolor{red}{\textbf{$22.670$}} \\
				\midrule  
				\multirow{3}{*}{Monte Carlo} 
				& $256 \times 256$  & $5.868$ & $0.735$ & $0.585$ & $0.545$ & $2.318$ \\
				& $512 \times 512$  & $26.577$ & $4.244$ & $2.878$ & $7.672$ & $9.536$ \\
				& $1024 \times 1024$ & \textcolor{red}{\textbf{$100.849$}} & \textcolor{red}{\textbf{$17.043$}} & \textcolor{red}{\textbf{$12.797$}} & \textcolor{red}{\textbf{$30.009$}} & \textcolor{red}{\textbf{$44.554$}} \\
				\bottomrule
			\end{tabular}
		}
	\end{table}
	
	Fig. \ref{fig:complex} corroborates the efficiency of WLC. The trends in the two charts are completely consistent. 
	As shown in Table \ref{tab:runtime_combined}, WLC is faster than AG. On real-world tests with $1024^2$ input size, WLC requires only $22.67$ ms, significantly outperforming AG ($70.55$ ms). While SD and EN are fast, previous experiments have shown that they are unusable. Furthermore, $22.67$ms corresponds to approximately $44$ FPS, which still meets the real-time requirements. Although SD achieves the lowest latency due to basic matrix operations, it lacks the semantic discriminability required for fusion evaluation.
	
	\subsubsection{Alignment with Machine Perception}
	
\begin{figure}[htbp]
		\centering
		\includegraphics[width=0.95\linewidth]{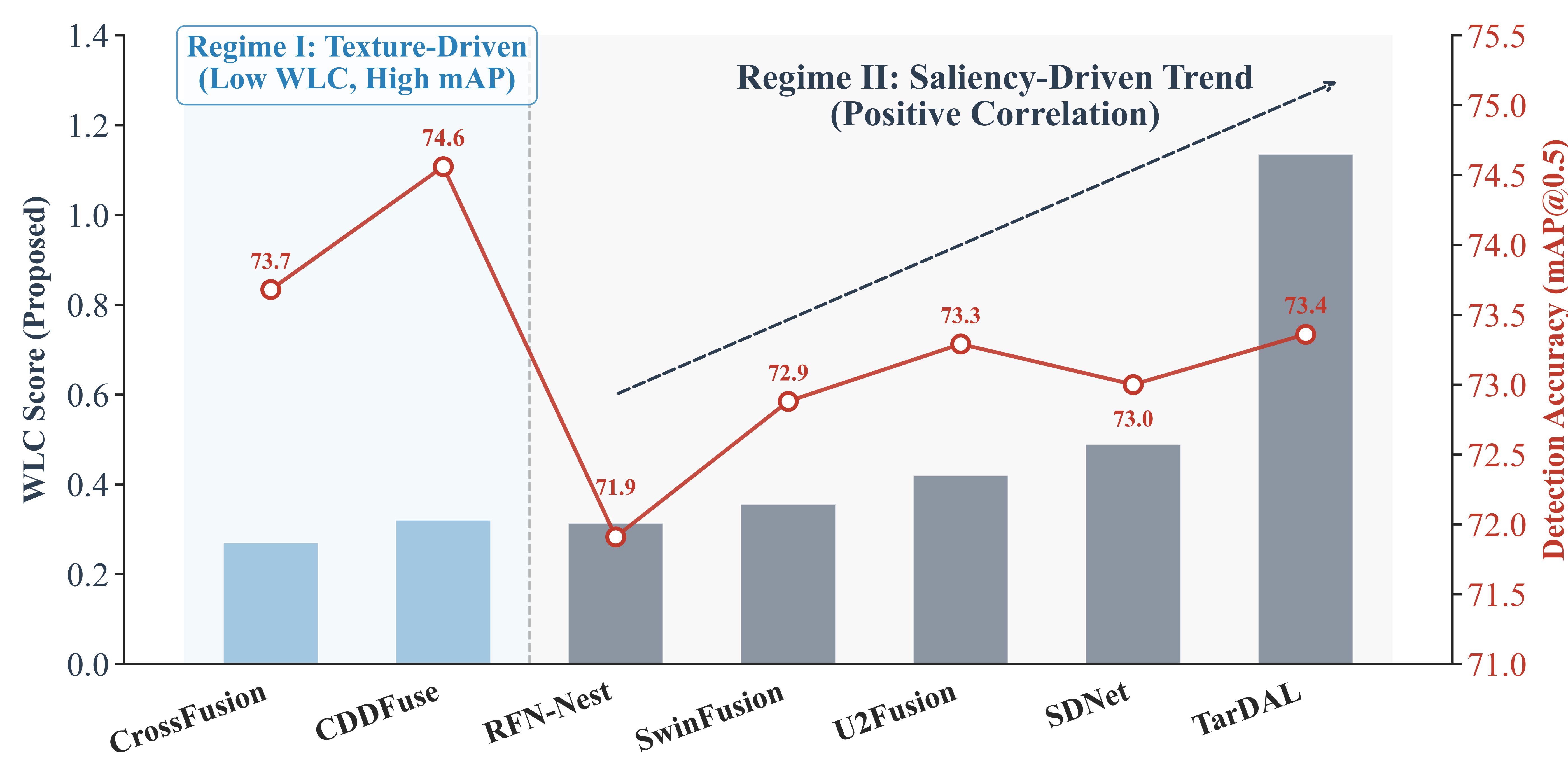}
		\caption{Correlation analysis between WLC scores and detection accuracy (mAP@0.5).
			The plot reveals a dual-mechanism relationship. 
			Regime I (Blue bars) represents texture-driven methods that achieve high accuracy through fine-grained feature preservation rather than contrast enhancement. 
			Regime II (Grey bars) demonstrates a strong positive correlation for saliency-driven methods, confirming that WLC directly boosts target visibility and detection recall. 
			The dashed arrow indicates the ideal optimization trajectory for saliency-based fusion frameworks.}
		\label{fig:map}
	\end{figure}

	Beyond perceptual quality, the ultimate benchmark for reconnaissance-oriented fusion lies in its utility for downstream machine perception tasks. To validate the practical value of WLC, this paper analyzes the correlation between metric scores and object detection performance (mAP@0.5) using the YOLOv8 detector on the DroneVehicle dataset. Rather than treating seven fusion algorithms (CrossFusion \cite{crossfuse}, CDDFusion \cite{cddfuse}, RFN-Nest \cite{rfn}, SwinFusion \cite{swinfusion}, U2Fusion \cite{u2fusion}, SDNet \cite{sdnet}, and TarDAL \cite{TarDAL}) as a homogeneous group, we categorize them based on their intrinsic optimization objectives documented in the literature. This mechanism-based perspective, visualized in Fig. \ref{fig:map}, reveals two distinct operating regimes. 
	
	This paper first identifies a ``Saliency-Driven Regime'' comprising methods that prioritize thermal intensity preservation, such as U2Fusion \cite{u2fusion}, SwinFusion \cite{swinfusion}, and TarDAL \cite{TarDAL}. For this group, we observe a distinct positive correlation between WLC scores and detection accuracy. Specifically, TarDAL, which explicitly incorporates a target-aware discriminator to maximize saliency, achieves both the highest WLC score ($1.13$) and superior detection performance within this group. Similarly, U2Fusion, designed to retain the maximum intensity from source images, also exhibits high values in both metrics. As WLC score increases from $0.31$ (RFN-Nest) to $1.13$ (TarDAL), the mAP consistently improves. This empirical trend substantiates the hypothesis that in low-altitude night scenarios, enhancing local thermal contrast serves as a prerequisite for the detector's Region Proposal Network (RPN) to successfully localize potential targets against background clutter. High WLC scores effectively predict the ``pop-out'' effect, acting as a reliable indicator for recall improvement. 
	
	Conversely, we observe a ``Texture-Restoration Regime'' represented by methods like CDDFuse \cite{cddfuse} and CrossFusion \cite{crossfuse}. These algorithms achieve competitive or even leading detection performance (e.g., $74.6\%$ for CDDFuse) despite exhibiting moderate WLC scores. We attribute this phenomenon to their underlying architectural focus: CDDFuse leverages Restormer blocks and invertible neural networks to maximize fine-grained detail recovery rather than simple contrast enhancement. Consequently, their success stems from the superior preservation of high-frequency visible features—such as edges and wheels—which facilitates the classification head of the detector. This indicates that while WLC ensures target visibility, texture fidelity guarantees identifiability. 
	
	In summary, we interpret WLC as a metric that quantifies the upper bound of target discoverability. It functions as a necessary precondition for detection in low-light conditions, complementing texture-based metrics to form a holistic evaluation of machine perception utility.

	\subsection{Limitations}
	Despite its robustness, WLC operates within specific theoretical boundaries governed by psychophysics and mathematical morphology.
	
	\subsubsection{Intensity Lower Bound}
	WLC metric is derived from Weber's Law, which posits that the contrast threshold $\Delta I$ is proportional to the background intensity $I$:
	\begin{equation}
		\frac{\Delta I}{I} = k \quad (\text{Weber Region})
	\end{equation}
	
	However, this linearity holds primarily in the photopic regime. In the scotopic (extreme low-light) regime, where $I$ approaches the sensor's noise floor, human perception transitions to the de Vries-Rose Law \cite{Rose}. In this regime, perception is limited by quantum fluctuations, and the threshold follows a square-root relationship:
	\begin{equation}
		\frac{\Delta I}{\sqrt{I}} = C \quad (\text{de Vries-Rose Region})
	\end{equation}
	
	As the background intensity $I \to 0$ (e.g., deep space or absolute darkness), the denominator in WLC formulation decays faster than the physical noise threshold $\sqrt{I}$, potentially leading to numerical instability. Although the smoothing term $\epsilon$ mitigates singularity, WLC is theoretically less reliable in the de Vries-Rose region compared to the Weber region. However, this does not mean we will choose the de Vries-Rose region. Since low-altitude reconnaissance primarily deals with suprathreshold targets, WLC is optimized for the Weber region to ensure semantic saliency measurement.
	
	\subsubsection{Spatial Lower Bound}
	WLC relies on morphological operations to define the local background, implicitly assuming the target is an extended source with a definable shape. 
	According to Ricco's Law \cite{Ricco}, for stimuli below a critical spatial area $A_{crit}$, perception is determined by total energy rather than contrast:
	\begin{equation}
		I \times A = C \quad (\text{for } A < A_{crit})
	\end{equation}
	
	This creates a spatial ``Blind Zone''. If the target size is smaller than the structuring element of the morphological dilation (e.g., $< 3 \times 3$ pixels), the background region cannot be topologically formed. Consequently, WLC is strictly valid for object-level targets rather than sub-pixel point sources.
	
	\subsubsection{Modality Dependency} 
	The formulation assumes a valid infrared prior. If the infrared saliency vanishes, the generated mask $\mathbf{M}_{obj}$ becomes empty, rendering the local contrast calculation undefined. This highlights that WLC is contingent on the physical visibility of thermal signatures.
	
	\section{Conclusion}
	\label{sec:conclusion}
	This paper identifies and addresses a critical evaluation paradox in low-altitude scenes, i.e., ``Noise Trap''. To resolve this, this paper introduces WLC. By shifting the evaluation paradigm from global pixel statistics to local semantic saliency, WLC establishes a robust, unsupervised standard that aligns with both human visual perception and machine detection utility.
	
	Mathematically, WLC functions as a ratio-based discriminator where the background mean intensity serves as a normalization factor. This structure inherently ensures noise robustness.Furthermore, the metric exhibits a ``target loss(\ref{target_loss})'' property. This is an excellent property that gradient-based metrics (AG and SF) and entropy-based metrics (EN and SD) cannot achieve. In terms of semantic discriminability, the statistical distribution of WLC scores on the DroneVehicle dataset confirms its sensitivity to real-world target diversity. While traditional metrics remain statistically ``numb'', producing low variance regardless of scene content, WLC captures the wide dynamic range of thermal signatures, ranging from scorching heat sources to subtle thermal traces. This sensitivity to local semantics allows WLC to distinguish between high-quality, target-preserving fusion and noise-corrupted results. Finally, from an algorithmic perspective, WLC maintains $O(N)$ linear computational complexity.

	Future work will focus on integrating WLC into the optimization process of deep learning frameworks to directly guide image enhancement. Additionally, we aim to extend this local contrast paradigm to other challenging fields with similar characteristics, further validating its potential in assisting diverse downstream machine perception tasks.
	In conclusion, WLC is not only a more accurate metric but also provides a new optimization direction for solving image fusion challenges in low-altitude environments.

	\balance
	\begin{IEEEbiography}[{\includegraphics[width=1in,height=1.25in,clip,keepaspectratio]{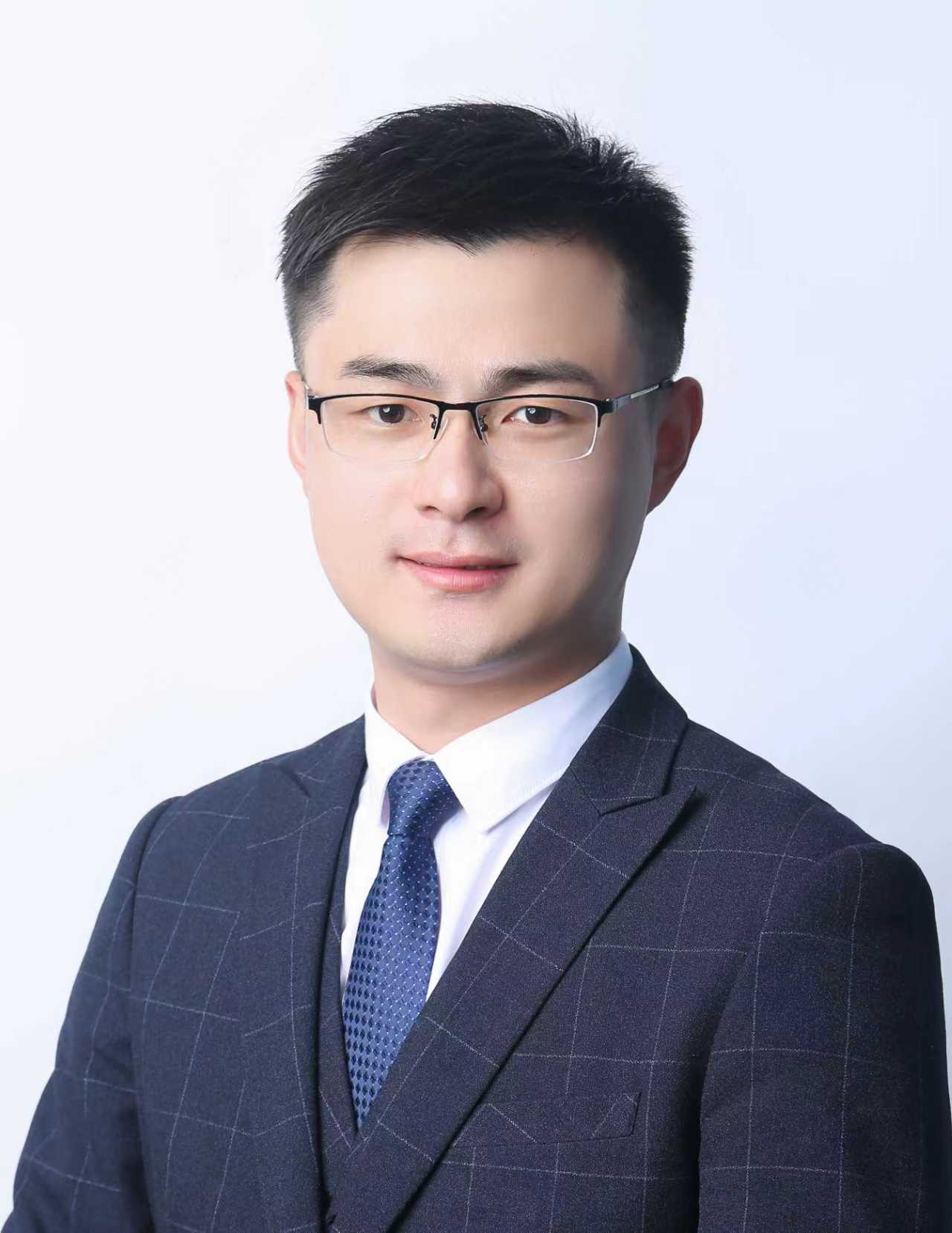}}]{Cong Wang}
		(Member, IEEE) received the B.S. degree in automation and the M.S. degree in mathematics from Hohai University, Nanjing, China, in 2014 and 2017, re-spectively, and the Ph.D. degree in mechatronic engineering from Xidian University, Xi'an, China, in 2021. He is currently an associate professor with the School of Mathematics and Statistics, Northwestern Polytechnical University, Xi'an. He was a visiting Ph.D. Student with the Department of Electrical and Computer Engineering, University of Alberta, Edmonton, AB, Canada, and the Department of Electrical and Computer Engineering, National University of Singapore, Singapore. He was also a research assistant with the School of Computer Science and Engineering, Nanyang Technological University, Singapore. His current research interests include vicinagearth security, high-dimensional image analysis, and fuzzy theory and its applications. He currently has a textbook, a monograph, 10+ patents and more than 50 papers (20+ in \textit{IEEE Transactions}) and hosts more than 10 research projects. He is funded by the China National Postdoctoral Program for Innovative Talents and the Excellent Chinese and Foreign Youth Exchange Program of the China Association for Science and Technology. He is an editorial board member of 10+ international journals like IEEE TFS and a program committee chair, track chair, publication chair, publicity chair,  program committee member, and technical committee member for 60+ international conferences. He is also a frequent reviewer of 70+ international journals, including a number of IEEE Transaction.
	\end{IEEEbiography}
	\vspace{-5mm}
	\begin{IEEEbiography}[{\includegraphics[width=1in,height=1.25in,clip,keepaspectratio]{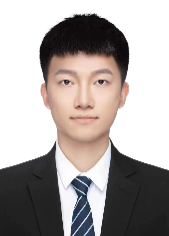}}]{Yufeng Xie}
		received the B.S. degree in mathematics from the Ocean University of China, Qingdao, China, in 2024. He is currently pursuing the M.S. degree in mathematics with the School of Mathematics and Statistics, Northwestern Polytechnical University, Xi'an, China. His research interests include multimodal image fusion.
	\end{IEEEbiography}
	
	\begin{IEEEbiography}[{\includegraphics[width=1in,height=1.25in,clip,keepaspectratio]{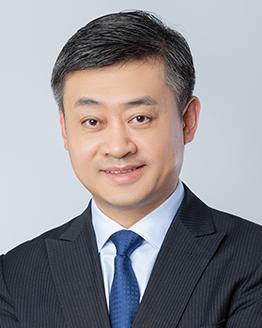}}]{XueLong Li}
		(Fellow, IEEE) is the CTO and Chief Scientist of China Telecom, where he founded the Institute of Artificial Intelligence (TeleAl) of China Telecom.
	\end{IEEEbiography}


\begin{thebibliography}{100}
	
	\bibitem{4ir}
	Y. Li, F. Wei, Y. Zhang, W. Chen, and J. Ma, ``HS2P: Hierarchical spectral and structure-preserving fusion network for multimodal remote sensing image cloud and shadow removal,'' \emph{Information Fusion}, vol. 94, pp. 215--228, 2023.
	
	\bibitem{5vis}
	X. Zhang, ``Deep learning-based multi-focus image fusion: A survey and a comparative study,'' \emph{IEEE Trans. Pattern Anal. Mach. Intell.}, vol. 44, no. 9, pp. 4819--4838, 2022.
	
	\bibitem{r1}
	H. Sun, Q. Liu, J. Wang, J. Ren, Y. Wu, H. Zhao, and H. Li, ``Fusion of infrared and visible images for remote detection of low-altitude slow-speed small targets,'' \emph{IEEE J. Sel. Topics Appl. Earth Observ. Remote Sens.}, vol. 14, pp. 2971--2983, 2021.
	
	\bibitem{r2}
	H. Fang, M. Xia, G. Zhou, Y. Chang, and L. Yan, ``Infrared small UAV target detection based on residual image prediction via global and local dilated residual networks,'' \emph{IEEE Geosci. Remote Sens. Lett.}, vol. 19, pp. 1--5, 2022.
	
	\bibitem{jc1}
	Y. Li, J. Luo, Y. Zhang, Y. Tan, J.-G. Yu, and S. Bai, ``Learning to holistically detect bridges from large-size VHR remote sensing imagery,'' \emph{IEEE Trans. Pattern Anal. Mach. Intell.}, vol. 46, no. 12, pp. 11507--11523, 2024.
	
	\bibitem{zz1}
	X. Zhang, P. Ye, H. Leung, K. Gong, and G. Xiao, ``Object fusion tracking based on visible and infrared images: A comprehensive review,'' \emph{Information Fusion}, vol. 63, pp. 166--187, 2020.
	
	\bibitem{ref20}
	L. Jiang, C. Wang, J. Yang, and X. Li, ``Feature-preserving fuzzy clustering for blurred G-image segmentation,'' \emph{IEEE Trans. Circuits Syst. Video Technol.}, to be published, doi: 10.1109/TCSVT.2026.3670307.
	
	\bibitem{ref26}
	H. Wang, C. Wang, and Y. Yuan, ``Hyperspectral image super-resolution via boundary perception and topology inference,'' \emph{IEEE Trans. Multimedia}, to be published, doi: 10.1109/TMM.2026.3668557.
	
	\bibitem{joint6}
	Z. Zhang, Y. Gao, M. Xiong, X. Luo, and X.-J. Wu, ``A joint convolution auto-encoder network for infrared and visible image fusion,'' \emph{Multimedia Tools and Applications}, vol. 82, no. 19, pp. 29017--29035, 2023.
	
	\bibitem{CNN7}
	X. Luo, J. Wang, Z. Zhang, and X.-J. Wu, ``A full-scale hierarchical encoder-decoder network with cascading edge-prior for infrared and visible image fusion,'' \emph{Pattern Recognition}, vol. 148, p. 110192, 2024.
	
	\bibitem{fusiongan}
	J. Ma, W. Yu, P. Liang, C. Li, and J. Jiang, ``FusionGAN: A generative adversarial network for infrared and visible image fusion,'' \emph{Information Fusion}, vol. 48, pp. 11--26, 2019.
	
	\bibitem{crossfuse}
	H. Li and X.-J. Wu, ``CrossFuse: A novel cross attention mechanism based infrared and visible image fusion approach,'' \emph{Information Fusion}, vol. 103, p. 102147, 2024.
	
	\bibitem{swinfusion}
	J. Ma, L. Tang, F. Fan, J. Huang, X. Mei, and Y. Ma, ``SwinFusion: Cross-domain long-range learning for general image fusion via swin transformer,'' \emph{IEEE/CAA J. Autom. Sinica}, vol. 9, no. 7, pp. 1200--1217, 2022.
	
	\bibitem{MemoryFusion}
	J. He, X. Luo, Z. Zhang, and X.-J. Wu, ``MemoryFusion: A novel architecture for infrared and visible image fusion based on memory unit,'' \emph{Pattern Recognition}, vol. 170, p. 112004, 2026.
	
	\bibitem{T2EA}
	Z. Huang, C. Lin, B. Xu, M. Xia, Q. Li, Y. Li, and N. Sang, ``T2EA: Target-aware Taylor expansion approximation network for infrared and visible image fusion,'' \emph{IEEE Trans. Circuits Syst. Video Technol.}, vol. 35, no. 5, pp. 4831--4845, 2025.
	
	\bibitem{FreqGAN}
	Z. Wang, Z. Zhang, W. Qi, F. Yang, and J. Xu, ``FreqGAN: Infrared and visible image fusion via unified frequency adversarial learning,'' \emph{IEEE Trans. Circuits Syst. Video Technol.}, vol. 35, no. 1, pp. 728--740, 2025.
	
	\bibitem{ref34}
	K. Li, D. Wang, H. Xu, H. Zhong, and C. Wang, ``Language-guided progressive attention for visual grounding in remote sensing images,'' \emph{IEEE Trans. Geosci. Remote Sens.}, vol. 62, pp. 1--13, 2024.
	
	\bibitem{ref28}
	C. Wang, J. Lin, and X. Li, ``Structural-equation-modeling-based indicator systems for image quality assessment,'' \emph{IEEE Trans. Pattern Anal. Mach. Intell.}, vol. 47, no. 8, pp. 7093--7107, 2025.
	
	\bibitem{SeAfuse}
	L. Tang, J. Yuan, and J. Ma, ``Image fusion in the loop of high-level vision tasks: A semantic-aware real-time infrared and visible image fusion network,'' \emph{Information Fusion}, vol. 82, pp. 28--42, 2022.
	
	\bibitem{TarDAL}
	J. Liu, X. Fan, Z. Huang, G. Wu, R. Liu, W. Zhong, and Z. Luo, ``Target-aware dual adversarial learning and a multi-scenario multi-modality benchmark to fuse infrared and visible for object detection,'' in \emph{Proc. IEEE/CVF Conf. Comput. Vis. Pattern Recognit. (CVPR)}, 2022, pp. 5792--5801.
	
	\bibitem{Lin2014COCO}
	T.-Y. Lin, M. Maire, S. Belongie, J. Hays, P. Perona, D. Ramanan, P. Doll\'{a}r, and C. L. Zitnick, ``Microsoft coco: Common objects in context,'' in \emph{Proc. Eur. Conf. Comput. Vis. (ECCV)}, 2014, pp. 740--755.
	
	\bibitem{target}
	H. Fang, M. Xia, G. Zhou, Y. Chang, and L. Yan, ``Infrared small UAV target detection based on residual image prediction via global and local dilated residual networks,'' \emph{IEEE Geosci. Remote Sens. Lett.}, vol. 19, pp. 1--5, 2022.
	
	\bibitem{weibo-Psychophysics}
	G. A. Gescheider, \emph{Psychophysics: The Fundamentals}, 3rd ed. Psychology Press, 1997.
	
	\bibitem{sun2020drone}
	Y. Sun, B. Cao, P. Zhu, and Q. Hu, ``Drone-based RGB-infrared cross-modality vehicle detection via uncertainty-aware learning,'' \emph{IEEE Trans. Circuits Syst. Video Technol.}, vol. 32, no. 10, pp. 6700--6713, 2022.
	
	\bibitem{Wave}
	H. Li, B. S. Manjunath, and S. K. Mitra, ``Multisensor image fusion using the wavelet transform,'' \emph{Graphical Models Image Process.}, vol. 57, no. 3, pp. 235--245, 1995.
	
	\bibitem{Pyramid}
	P. J. Burt and E. H. Adelson, ``The Laplacian pyramid as a compact image code,'' \emph{IEEE Trans. Commun.}, vol. 31, no. 4, pp. 532--540, 1983.
	
	\bibitem{densefuse}
	H. Li and X.-J. Wu, ``DenseFuse: A fusion approach to infrared and visible images,'' \emph{IEEE Trans. Image Process.}, vol. 28, no. 5, pp. 2614--2623, 2019.
	
	\bibitem{mamba}
	X. Xie, Y. Cui, T. Tan, X. Zheng, and Z. Yu, ``Fusionmamba: Dynamic feature enhancement for multimodal image fusion with mamba,'' \emph{Visual Intelligence}, vol. 2, no. 37, 2024.
	
	\bibitem{deep}
	L. Tang, H. Zhang, H. Xu, and J. Ma, ``Deep learning-based image fusion: a survey,'' \emph{J. Image Graph.}, vol. 28, no. 1, pp. 3--36, 2023.
	
	\bibitem{EN}
	J. W. Roberts, J. A. van Aardt, and F. B. Ahmed, ``Assessment of image fusion procedures using entropy, image quality, and multispectral classification,'' \emph{J. Appl. Remote Sens.}, vol. 2, no. 1, p. 023522, 2008.
	
	\bibitem{MI}
	C. E. Shannon, ``A mathematical theory of communication,'' \emph{Bell Syst. Tech. J.}, vol. 27, no. 3, pp. 379--423, 1948.
	
	\bibitem{SF}
	A. M. Eskicioglu and P. Fisher, ``Image quality measures and their performance,'' \emph{IEEE Trans. Commun.}, vol. 43, no. 12, pp. 2959--2965, 1995.
	
	\bibitem{AG}
	G. Cui, H. Feng, Z. Xu, Q. Li, and Y. Chen, ``Detail preserved fusion of visible and infrared images using regional saliency extraction and multi-scale image decomposition,'' \emph{Optics Commun.}, vol. 341, pp. 199--209, 2015.
	
	\bibitem{VIF}
	Y. Han, Y. Cai, Y. Cao, and X. Xu, ``A new image fusion performance metric based on visual information fidelity,'' \emph{Information Fusion}, vol. 14, no. 2, pp. 127--135, 2013.
	
	\bibitem{congwang}
	C. Wang, X. Tan, X. Ren, and X. Li, ``Fast and robust strain signal processing for aircraft structural health monitoring,'' \emph{J. Autom. Intell.}, vol. 3, no. 3, pp. 160--168, 2024.
	
	\bibitem{Digital}
	R. C. Gonzalez and R. E. Woods, \emph{Digital Image Processing}, 3rd ed. Upper Saddle River, NJ: Prentice Hall, 2008.
	
	\bibitem{Probability}
	E. T. Jaynes and G. L. Bretthorst, \emph{Probability Theory: The Logic of Science}. Cambridge, UK: Cambridge Univ. Press, 2003.
	
	\bibitem{Raliy}
	A. Papoulis and H. Saunders, \emph{Probability, Random Variables and Stochastic Processes}, 2nd ed. New York, NY: McGraw-Hill, 1989.
	
	\bibitem{segment-weibo}
	J. Chen, G. Zhao, M. Salo, E. Rahtu, and M. Pietikainen, ``Automatic dynamic texture segmentation using local descriptors and optical flow,'' \emph{IEEE Trans. Image Process.}, vol. 22, no. 1, pp. 326--339, 2013.
	
	\bibitem{face-weibo}
	J. Xu, ``Enhanced local face feature encoding via the Petersen graph framework,'' in \emph{Proc. 5th Int. Conf. Comput. Eng. Intell. Control (ICCEIC)}, 2024, pp. 169--172.
	
	\bibitem{brain-weibo}
	M. Punitha and M. S. Kumar, ``Core tumor prediction and feature segmentation by using Weber’s law in brain images,'' in \emph{Proc. 8th Int. Conf. I-SMAC}, 2024, pp. 1224--1228.
	
	\bibitem{feedback-weibo}
	X. Meng, J. Meng, G. Chai, X. Sheng, and X. Zhu, ``Discrete tactile feedback based on Weber’s law enhances prosthetic hand approaching performance under divided visual attention,'' \emph{IEEE Trans. Neural Syst. Rehabil. Eng.}, vol. 34, pp. 674--685, 2026.
	
	\bibitem{echo-weibo}
	H. Mehdia, M. A. Zakaria, D. Fethi, M. A. Abdelhak, B. Hadjira, R. Abdenebi, N. Abdelkrim, and M. Omar, ``Advancing echo processing techniques by combining support vector machine and Weber's law integration,'' in \emph{Proc. 3rd Int. Conf. Adv. Electr. Eng. (ICAEE)}, 2024, pp. 1--6.
	
	\bibitem{ref50}
	C. Wang, Z. Yan, W. Pedrycz, M. Zhou, and Z. Li, ``A weighted fidelity and regularization-based method for mixed or unknown noise removal from images on graphs,'' \emph{IEEE Trans. Image Process.}, vol. 29, pp. 5229--5243, 2020.
	
	\bibitem{ref22}
	J. Jing, S. Qu, C. Wang, and X. Li, ``Adaptively weighted residual-driven fuzzy C-means for image segmentation in presence of mixed or unknown noise,'' \emph{IEEE Trans. Fuzzy Syst.}, to be published, doi: 10.1109/TFUZZ.2026.3657734.
	
	\bibitem{TNO}
	A. Toet, ``The TNO multiband image data collection,'' \emph{Data in Brief}, vol. 15, pp. 249--251, 2017.
	
	\bibitem{M3FD}
	J. Liu, X. Fan, Z. Huang, G. Wu, R. Liu, W. Zhong, and Z. Luo, ``Target-aware dual adversarial learning and a multi-scenario multi-modality benchmark to fuse infrared and visible for object detection,'' in \emph{Proc. IEEE/CVF Conf. Comput. Vis. Pattern Recognit. (CVPR)}, 2022, pp. 5792--5801.
	
	\bibitem{cddfuse}
	Z. Zhao, H. Bai, J. Zhang, Y. Zhang, S. Xu, Z. Lin, R. Timofte, and L. Van Gool, ``Cddfuse: Correlation-driven dual-branch feature decomposition for multi-modality image fusion,'' in \emph{Proc. IEEE/CVF Conf. Comput. Vis. Pattern Recognit. (CVPR)}, 2023, pp. 5906--5916.
	
	\bibitem{rfn}
	H. Li, X.-J. Wu, and J. Kittler, ``RFN-Nest: An end-to-end residual fusion network for infrared and visible images,'' \emph{Information Fusion}, vol. 73, pp. 72--86, 2021.
	
	\bibitem{u2fusion}
	H. Xu, J. Ma, J. Jiang, X. Guo, and H. Ling, ``U2Fusion: A unified unsupervised image fusion network,'' \emph{IEEE Trans. Pattern Anal. Mach. Intell.}, vol. 44, no. 1, pp. 502--518, 2022.
	
	\bibitem{sdnet}
	H. Zhang and J. Ma, ``SDNet: A versatile squeeze-and-decomposition network for real-time image fusion,'' \emph{Int. J. Comput. Vis.}, vol. 129, no. 10, pp. 2761--2785, 2021.
	
	\bibitem{Rose}
	A. Rose, ``The sensitivity performance of the human eye on an absolute scale,'' \emph{J. Opt. Soc. Am.}, vol. 38, no. 2, pp. 196--208, 1948.
	
	\bibitem{Ricco}
	S. H. Schwartz, \emph{Visual Perception: A Clinical Orientation}, 4th ed. Medical, 2009.
	
\end{thebibliography}
\end{document}